\documentclass[11pt]{article}
\usepackage{fullpage}

\usepackage{booktabs} 
\usepackage{multirow}
\usepackage{amsmath, amsthm, amssymb, amsfonts}
\usepackage[truedimen,margin=25truemm]{geometry}

\usepackage{graphicx} 
\usepackage{subfigure} 

\usepackage{algorithm}
\usepackage{algorithmic}

\usepackage{color}

\usepackage{enumitem}

\usepackage{url}

\date{ }

\begin{document}

\title{Detection of Review Abuse via Semi-Supervised Binary Multi-Target Tensor Decomposition}

\author{Anil R. Yelundur\thanks{India ML - Amazon (e-mail: yelundur@amazon.com).} \and Vineet Chaoji\thanks{India ML - Amazon (e-mail: vchaoji@amazon.com).} \and Bamdev Mishra\thanks{Microsoft India (e-mail: bamdevm@microsoft.com).}}

\maketitle

\begin{abstract}
Product reviews and ratings on e-commerce websites provide customers with detailed insights about various aspects of the product such as quality, usefulness, etc. Since they influence customers' buying decisions, product reviews have become a fertile ground for abuse by sellers (colluding  with reviewers) to promote their own products or to tarnish the reputation of competitor's products. In this paper, our focus is on detecting such abusive entities (both sellers and reviewers) by applying tensor decomposition on the product reviews data. While tensor decomposition is mostly unsupervised, we formulate our problem as a semi-supervised binary multi-target tensor decomposition, to take advantage of currently known abusive entities. We empirically show that our multi-target semi-supervised model achieves higher precision and recall in detecting abusive entities as compared to unsupervised techniques. Finally, we show that our proposed stochastic partial natural gradient inference for our model empirically achieves faster convergence than stochastic gradient and Online-EM with sufficient statistics.
\end{abstract}

%
%


\section{Introduction}
Product reviews and ratings on e-commerce websites provide customers with detailed insights about various aspects of the product. Ratings allow customers to gauge the quality of the product as perceived by other customers who have bought the product. Consequently, customers rely significantly on product reviews and ratings while making buying decisions on e-commerce platforms. Given their influence on customer spends, product reviews are a fertile ground for abuse. 

\begin{figure}[t]
\center
\includegraphics[scale=0.50]{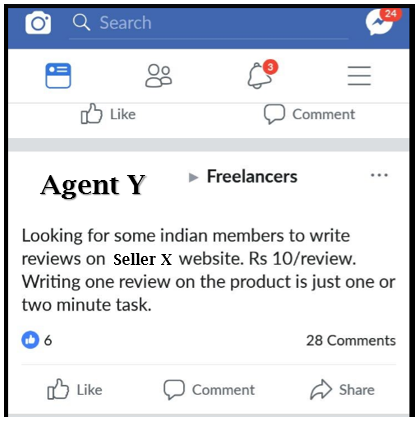}
\caption{\label{FB1:fig} Seller/agency soliciting fake reviewers.}
\end{figure}

A common form of abuse involves a seller running campaigns soliciting fake, genuine-looking positive reviews, for their own products or fake negative reviews about their competitors' products. The paid reviewers have a varied modi operandi. They can either create their own account and start posting paid reviews (fake reviews) or they can hijack an inactive account in good standing to post seemingly innocuous reviews from that account. There are also businesses/agencies which promise a fee for writing fake reviews on Amazon. Figure~\ref{FB1:fig} shows a social media snippet (name anonymized) of a seller or an agency soliciting reviewers for a fee. To circumvent identification, paid reviewers also distribute the volume of fake reviews across multiple accounts.

In order to maintain customer's trust on the reviews and in turn on the e-commerce platform, it is imperative for e-commerce websites to ensure that the reviews remain sacrosanct. As a result, identifying and taking enforcement actions on fake reviews, paid reviewers and the underlying abusive sellers is a significant focus area within e-commerce companies. 
In this paper, we broadly focus on the problem of detecting abusive entities (sellers and reviewers) within the product reviews' ecosystem. We formalize the interactions between sellers, products and reviewers as a binary tensor. Tensors are multidimensional arrays \cite{kolda2009tensor} and are used in capturing multidimensional features effectively. We subsequently apply tensor decomposition techniques to identify the dense cores. These dense cores are treated as anomalous interactions within a tensor with predominantly uniformly distributed interactions. Additionally, we use known abusive sellers and reviewers from the past as partial supervision to further enhance the model, with each form of abuse as a separate target signal. To summarize, our technical contributions are as follows:
\begin{enumerate}
\item We formulate detection of abusive entities (sellers and reviewers), a task of identifying dense cores in the seller-reviewer bipartite graph, as a tensor decomposition problem.
\item We apply unsupervised Bayesian binary tensor decomposition to detect dense blocks in the seller-reviewer relationship. This is based on the Logistic CP tensor decomposition model.
\item We then develop {\bf se}mi-supervised bi{\bf n}ary mul{\bf ti}-target extension to the u{\bf n}supervised mod{\bf el} (via P\'{o}lya-Gamma data augmentation) so that we can incorporate prior information about multiple forms of abuse to improve detection of abusive entities. We call our proposed approach {\bf {\em SENTINEL}}. 
\item Finally, we develop stochastic partial natural gradient learning for the semi-supervised model and show that it empirically achieves faster convergence than stochastic gradient descent and EM with sufficient statistics.
\item We show the efficacy of the proposed approach as compared to the state-of-the-art techniques for tensor decomposition and review abuse detection.
\end{enumerate}
To the best of our knowledge, this is the first time a) semi-supervised multi-target binary tensor decomposition has been applied to the problem of detecting fake entities in the review spam domain, b) natural gradients has been used for inference within tensor decomposition.
Although the paper focuses on the problem formulation and scientific aspects of SENTINEL, we have additionally developed an extensive platform that uses machine learning models to flag abusive entities. The platform allows a combination of automated as well as manual enforcement. The feedback from the enforcement gets channeled back into the platform to further enhance SENTINEL.

The rest of this paper is organized as follows. Section~\ref{sec:related_work} introduces related works as well as some background regarding our application of tensor decomposition for detecting abuse in the seller-reviewer relationship. Section \ref{sec:un_super} describes the baseline unsupervised binary tensor decomposition technique. Section \ref{sec:multi_target} describes SENTINEL, encapsulating the semi-supervised multi-target extensions. Section \ref{sec:Mstep} describes our proposed stochastic partial natural gradient learning for the inference of all the latent parameters of the semi-supervised model. Experimental results are shown in Section \ref{sec:experiments}.

\section{Related Work and Background}
\label{sec:related_work}
There has been a lot of attention recently to address the issue of finding fake reviewers in online e-commerce platforms. Jindal et al.~\cite{jindal07} were one of the first to show that review spam exists and proposed simple text based features to classify fake reviewers.

{\bf Identifying Abusive Reviews:} Abusive reviewers have grown in sophistication ever since the initial efforts~\cite{jindal07,jindal08a}, employing professional writing skills to avoid detection via text-based techniques. In~\cite{feng12}, the authors have proposed stylistic features derived from the Probabilistic Context Free Grammar parse trees to detect review spam. To detect more complex fake review patterns, researchers have proposed 1) graph based approaches such as approximate bipartite cores and lockstep behavior detection among reviewers \cite{li16a, beutal13a, hooi16a, jiang15a}, 2) techniques to identify network footprints of reviewers in the reviewer product graph \cite{ye15a}, and 3) using anomalies in ratings distribution \cite{hooi16b}. Some recent research has pointed at the importance of time in identifying fake reviews since it is critical to produce as many reviews as possible in a short period of time to be economically viable. Methods exploiting temporal and spatial features related to reviewers/reviews~\cite{li15a,ye16a}, as well as the sequence of reviews~\cite{lin14} have been proposed. While it is not possible to capture all the work on review spam detection,~\cite{Crawford2015} provides a broad coverage of efforts in this area.

{\bf Tensor based methods:} Techniques such as CrossSpot \cite{jiang15b}, M-Zoom \cite{shin16a}, and MultiAspectForensics \cite{maruhashi11a} propose identifying dense blocks in tensors or dense sub-graphs in heterogeneous networks, which can also be applied to the problem of identifying fake reviewers. 
M-Zoom is an improved version of CrossSpot that computes dense blocks in tensors which indicate anomalous or fraudulent behavior. The number of dense blocks (i.e., sub-tensors) returned by M-Zoom is configured a-priori. Note that the dense blocks identified may be overlapping, i.e., a tuple could be included in two or more blocks. M-Zoom is used as one of the baseline unsupervised methods in our experiments. MultiAspectForensics, on the other hand, automatically detects and visualizes novel patterns that include bipartite cores in heterogeneous networks.

\begin{figure}[t]
\center
\includegraphics[scale=0.5]{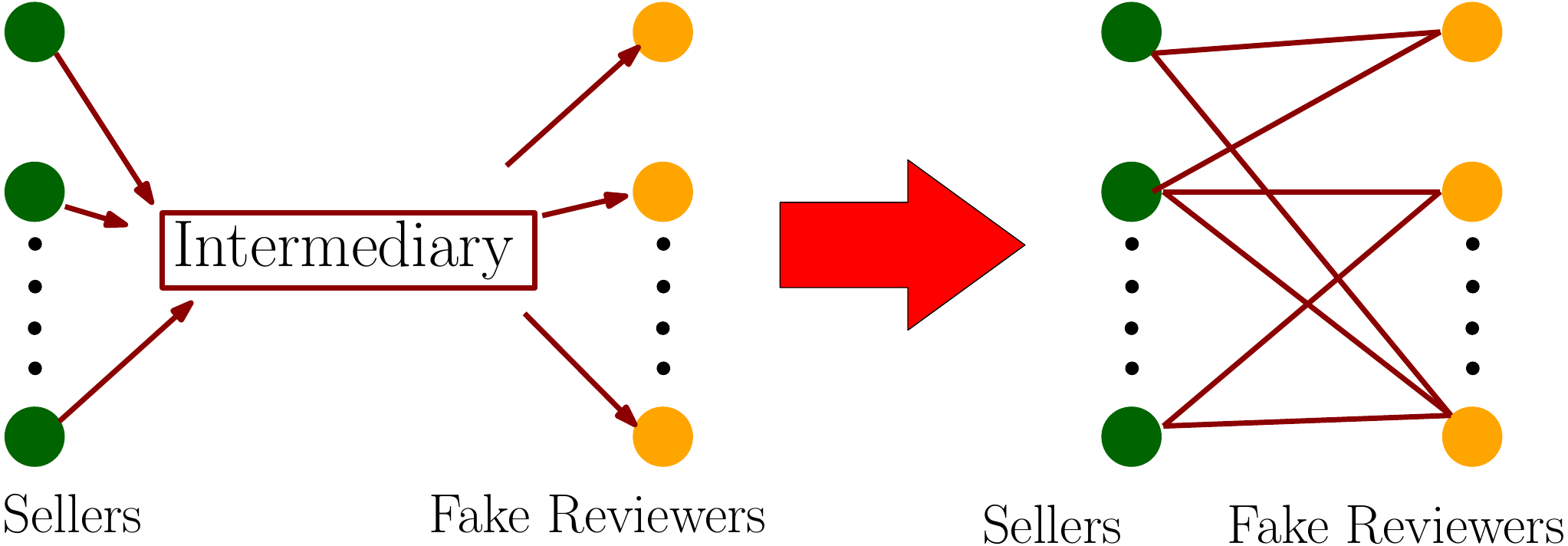}
\caption{\label{biPart:fig} Seller-Reviewer Abuse: key signals.}
\end{figure}

Tensor decomposition \cite{schein15a, hu15a, rai14a, rai15a, rai15b} is also applied to detect abusive entities (i.e., sellers and reviewers) since it facilitates the detection of dense bipartite sub-graphs. Dense bipartite sub-graphs are indicative of suspicious behavior due to the following reason: at any given time, there is a common pool of fake reviewers (paid reviewers) that are available and who are willing to write a positive or negative review in exchange for a fee. Sellers recruit these fake reviewers through an intermediary channel (such as social media groups, third party brokers, etc.) as shown in Figure~\ref{biPart:fig}, where nodes on the left indicate sellers and nodes on the right indicate reviewers and an edge indicates a written review. Note that a seller has to recruit a \textit{sizeable} number of fake reviewers in a short amount of time to make an impact on the overall product rating -- positive impact for her own products and negative impact for her competitor's products. Given a common pool of available fake reviewers at any given time, suspicious behavior manifests as dense bipartite connections between a group of sellers and a group of reviewers with similar ratings, such as near $5$-star or near $1$-star ratings. Hence the presence of a bipartite core (a dense bipartite sub-graph) in some contiguous time interval $\Delta t$ is a strong indicator of abuse by the group of sellers and reviewers involved.

Therefore, our goal boils down to finding bipartite cores (or dense blocks) using the factor matrices resulting from tensor decomposition. The modes of the tensor for our problem space correspond to the seller, reviewer, product, rating, and time of review. The entities in the corresponding factor matrices that have relatively higher values are indicative of abuse. By aggregating these abusive entities across the modes of the tensor results in discovering bipartite cores, where each core consists of a group of reviewers that provide similar rating across a similar set of sellers (and their products) where all of these reviews are occurring within a short contiguous interval of time.

We have a small set of known abusive sellers and reviewers identified via manual audits. We leverage the partial supervision to propose a semi-supervised extension to tensor decomposition to detect new abusive entities with greater fidelity. To leverage correlations between different forms of abuse (e.g., paid reviews abuse, abuse related to compromised accounts), we incorporate multiple binary targets based on Logistic Model with P\'{o}lya-Gamma data augmentation. 

{\bf Natural gradient learning:} Natural gradient learning~\cite{amari98a} is an alternative to traditional gradient descent based learning. We develop stochastic partial natural gradient learning for the semi-supervised tensor decomposition model and show that it empirically achieves faster convergence as compared to stochastic gradient descent and EM with sufficient statistics.

\section{Logistic CP Decomposition}
\label{sec:un_super}
Let ${\mathcal X}$ be a 3-mode (aka 3-way) tensor. CP decomposition of a tensor is defined as:
\begin{equation} \label{tensorD:eqn}
{\mathcal X} = \sum_{r=1}^R \lambda_r \vec{a}_r \odot \vec{b}_r \odot \vec{c}_r , 
\end{equation}
where $\lambda_r$ denotes the weight, $\vec{a}_r$, $\vec{b}_r$, and $\vec{c}_r$ are vectors (or rank-1 tensors) and $\odot$ represents vector outer product. $R$ is called the rank of the tensor. \textit{CP Decomposition} is a generalization of matrix decomposition.
Each tuple in Amazon's review data captures the `reviewing/rating' event wherein a reviewer writes a review/rating at a specific time for a product sold by a seller. These relationships can be described as a 5-way (or 4-way, if the seller is omitted) binary tensor. Such a binary tensor is incomplete since not all reviewers would have rated all products. Hence, the elements of a binary tensor can be modeled as being generated from a Logistic function, i.e., a value of \textit{1} when the relationship exists and value of \textit{0} where relationships do not exist. In our probabilistic tensor decomposition framework, we only consider those entries where a current relationship exists.

Logistic CP tensor decomposition is tensor CP decomposition in a probabilistic framework \cite{rai14a, rai15a, rai15b}. The generative model for a $K$-mode tensor denoted by ${\mathcal X}$ is:
\begin{equation*}
\begin{array}{lllll}
{\mathcal X} \sim f(\sum_{r=1}^R \lambda_r \vec{u}^{(1)}_r \odot \cdots \odot \vec{u}^{(K)}_r), \\
\delta_l \sim \mbox{Inv-Gamma}(a_l, 1): \mbox{$a_1 = 1, a_l = a_1 + (l - 1)\frac{1}{R}$}, \\ 
\tau_r = \prod_{l=1}^{r} \delta_l, \\
\lambda_r \sim \mathcal{N}(0, \tau_r), \\
\boxed{\mu_{i_k,r}^{(k)} \sim \mbox{Inv-Gamma}(a_c, b_2): \mbox{\small{\textbf{Our Enhancement}}},} \\
u_{i_k,r}^{(k)} \sim \mathcal{N}(0, \mu_{i_k,r}^{(k)}), \\
\end{array}
\end{equation*}
where $f$ specifies the Bernoulli-Logistic function for the binary valued tensor.
Let $\vec{u}^{(k)}_{i_k}$ denote the $R$ dimensional factor corresponding to entity $i_k$ in mode $k$. The number of non-zero values of $\vec{\lambda}$ determines the \textit{rank} of the tensor. Gaussian priors are assigned to the latent variables $\vec{\lambda}$ and $\vec{u}^{(k)}_{i_k}$ for $k \in [1, K]$. The variance of the Gaussian prior for $\vec{\lambda}$ is controlled by a \textit{Multiplicative Inverse Gamma Process} \cite{durante16a} that has the property of reducing the variance as $r$ increases. To get closed-form updates for all the latent variables, \cite{rai14a, rai15a, rai15b} introduce additional variables, denoted by $\omega_i$ corresponding to each tuple $i$ in the binary tensor that are P\'{o}lya-Gamma distributed.

We assign an Inverse-Gamma hyper-prior to the variance of the Gaussian priors of $\vec{u}^{(k)}_{i_k}$. The reason for adding the Inverse-Gamma hyper-prior to the model is that it provides adaptive L2-regularization. Hence, we can control the amount of regularization, i.e., provide different amounts of regularization to factors of the mode(s) that have target information than those without target information.

We use this unsupervised model as our base and develop multi-target semi-supervised extensions to it as described in Section \ref{sec:multi_target}. \cite{rai14a, rai15a, rai15b} propose using either sufficient statistics (in batch EM or as online EM) or Gibbs sampling (in batch) for inference. They claim that the online EM reaches reasonably good quality solutions fairly quickly as compared to their batch counterparts in most situations. However, the online EM does scalar updates of each latent variable that is inherently a vector of dimension $R$, and this may result in slower convergence. Alternatively, we propose using partial natural gradient learning in a stochastic setting for inference that is both, online in nature as well as allows vectorized updates for each of the latent variables. 


\section{Proposed Multi-Target Semi-Supervised Logistic CP Decomposition}
\label{sec:multi_target}
In this section, we describe SENTINEL, the semi-supervised extensions to the Logistic CP model for detecting abusive sellers and reviewers. We have prior information associated with a subset of the entities for at least one of the modes. This prior information is specified as binary target(s), where each target corresponds to a specific type of abuse. The framework is called \textit{semi-supervised tensor decomposition} as the tensor decomposition is achieved by incorporating this prior data. The intuition behind using the target information is that the patterns hidden in the known abusive entities could be leveraged to discover, with greater precision, additional entities that have similar signatures. We know that abusive behavioral patterns change as the entities continuously try to game the system. We hypothesize that the proposed semi-supervised framework offers a way to detect such changing behavior earlier and with greater efficiency as compared with unsupervised techniques.

In multi-target semi-supervised tensor decomposition, one or more targets are specified for at least one of the modes (denoted as $\vec{z}_l^{(k)}$ for target $l$ in mode $k$). We drop the superscript $(k)$ for convenience. This is represented as a matrix where each column corresponds to a single target. For each target:
\newline
1)~~~~~Both positive and negative labels need to be specified.
\newline
2)~~~~~Data can be specified for only a subset of the entities in that mode (semi-supervised learning).

The decomposition of the tensor is achieved by taking the multi-target label information across the mode(s) into account. The label information corresponding to each target is predicted via its own Logistic model whose coefficients are denoted as $\vec{\beta}_l^{(k)}$ for target $l$ in mode $k$. We drop the superscript $(k)$ for convenience. For a given mode, the covariates of each Logistic model are the same, i.e., the factors corresponding to the entities in that mode. The coefficients $\vec{\beta}_l$ are assigned \textit{Gaussian} priors. Note that the Logistic model corresponding to each target is learnt by taking into account the correlations of the binary labels from other targets, and therefore called as \textit{Multi-target Learning}. For each Logistic model, to get closed form updates, we introduce additional variables denoted by $\nu_l$ that are \textit{P\'{o}lya-Gamma} distributed. Equation \ref{fNuPG:eqn} in Appendix (Section \ref{sec:detBeta}) shows how the Logistic likelihood for each task $l$ becomes a quadratic function in $\vec{\beta}_l$ by augmenting the data with $\nu_l$.

\subsection*{Multi-Target Learning}
Consider binary target information for mode $k$ that is specified as an $M{\times}L$ matrix, where $M$ denotes the number of entities for which the targets are specified and $L$ denotes the number of tasks.
Note that in the semi-supervised setting the binary target information is usually specified for only a subset of the entities of some mode $k$. Based on the specified target information, the tensor decomposition technique can infer the neighbors of these abusive entities. For example, if a seller $a$ is flagged as having review abuse, then the tensor decomposition technique would infer its factor as a function of the reviewers associated with this seller during some time $\Delta t$ where the density is high. This in turn would lead to detection of other sellers who are also connected to some subset of these same reviewers during this same time interval. These other sellers would then end up having a similar factor representation indicating a high probability of being abusive.

Given $L$ tasks (where $L > 1$), the learning falls under the multi-target learning framework. Multi-target learning takes into account the correlations (or a similarities) between the $L$ tasks to learn $L$ different classifiers \cite{huang15a}. This is achieved via defining a $L{\times}L$ matrix where each element $l,j$ of that matrix, denoted by $Q_{l,j}$, carry the reverse cosine information between the targets $l$ and $j$ and is computed as:
\begin{equation*}
Q_{l,j} = 1 - \frac{\vec{z}_l^{\top}\vec{z}_j}{|\vec{z}_l|.|\vec{z}_j|}. \\
\end{equation*}
The interpretation of this $L{\times}L$ matrix denoted by $\boldsymbol{Q}$ is easily understood by taking a single task $l$ as an example. Note that all the diagonal elements of $\boldsymbol{Q}$ are zero (by definition). For task $l$, compute the following quantity given below:
\begin{equation}
e^{-\sum_{j \neq l} Q_{l,j}(\vec{\beta_{l}}^{\top}\vec{\beta_{j}})}.
\label{mltTgt:eqn}
\end{equation}

Equation \ref{mltTgt:eqn} can be regarded as an exponential distribution on $\vec{\beta_{l}}$ (without the normalizing factor) with a rate parameter equal to $\sum_{j \neq l} Q_{l,j}.\vec{\beta_{j}}$. We want to find a suitable $\vec{\beta_{l}}$ given the rate parameter such that it maximizes this exponential distribution. This is described considering the three extreme cases:
\newline
1) If target $l,j$ are negatively-correlated (cosine measure near $-1$ and $Q_{l,j} \sim 2$), then the coefficients defining their respective classifiers need to be of opposite signs and similar magnitude. This implies that the dot product $\vec{\beta_{l}}^{\top}\vec{\beta_{j}}$ is negative. And since $Q_{l,j}$ is positive, implies that the exponent is positive.
\newline
2) If target $l,j$ are uncorrelated (cosine measure near $0$ and $Q_{l,j} \sim 1$), then the coefficients defining their respective classifiers need to be distinct as well. This implies that the dot product $\vec{\beta_{l}}^{\top}\vec{\beta_{j}}$ will be closer to zero, meaning that the exponent is close to zero.
\newline
3) If target $l,j$ are highly correlated (cosine measure near $1$ and $Q_{l,j} \sim 0$), then the coefficients defining their respective classifiers need to be very similar (same sign and similar magnitude). This implies that the dot product $\vec{\beta_{l}}^{\top}\vec{\beta_{j}}$ will be positive. However, since $Q_{l,j} \sim 0$, implies that the exponent will be close to zero.

To handle cases when the magnitude any of element(s) $\vec{\beta_{j}}$ are very small or very large, we modify equation \ref{mltTgt:eqn} as given below:
\begin{equation}
e^{-\sum_{j \neq l} Q_{l,j}[\vec{\beta_{l}}^{\top}Sign(\vec{\beta_{j}})]}.
\label{mltTgt_1:eqn}
\end{equation}
Taking the natural logarithm of equation \ref{mltTgt_1:eqn} yields:
\begin{equation*} -\sum_{j \neq l} Q_{l,j}[\vec{\beta_{l}}^{\top}Sign(\vec{\beta_{j}})]. \end{equation*}
The Appendix (sections \ref{sec:detBeta} and \ref{sec:detUfactor}) describes the modeling of the $L$ classifier models in a multi-target framework jointly with the factors from the Logistic CP decomposition. Section \ref{sec:detLam} in the Appendix describes the modeling of the latent parameter $\vec{\lambda}$. Next section describes our proposed partial natural gradient inference for the multi-target semi-supervised Logistic CP model.

\section{Partial Natural Gradients: Inference}
\label{sec:Mstep}

Natural gradient is defined as the product of the Euclidean, i.e., standard gradient and the inverse of the Fisher information matrix. Natural gradient learning in the context of online learning is explained in \cite{amari98a}. It is an optimization method that is traditionally motivated from the perspective of information geometry and works well for many applications as an alternate to stochastic gradient descent \cite{martens14a}. Natural gradient descent is generally applicable to the optimization of probabilistic models. It has been shown that in many applications, natural gradients seem to require far fewer total iterations than gradient descent, hence making it a potentially attractive alternate method. However it has been known that for models with many parameters, computing the natural gradient is impractical since it requires computing the inverse of a large matrix, i.e., the Fisher information matrix. This problem has been addressed in prior works where an approximation to the Fisher is calculated such that it is easier to store and invert than the exact Fisher.

The latent variables in our semi-supervised model that we need to infer are $\vec{\lambda}$ of length $R$, matrix $\boldsymbol{U}^{(k)}$ for each mode $k \in [1, K]$ of dimension $n_k \times R$ and $\vec{\beta}_l^{(k)}$ of length $R+1$ for target $l \in [1:L]$ in mode $k \in [1, K]$. The corresponding log-posteriors of these latent variables that we need to maximize are denoted by $\log[g(\vec{\lambda})]$ (defined in \ref{sec:detLam}), $H(\vec{u}_{i_k=n}^{(k)})$ (defined in \ref{sec:detUfactor}) and $F(\vec{\beta}_l^{(k)})$ (defined in \ref{sec:detBeta}), respectively. Natural gradient update in iteration $t$ is then defined as:
{\small
\begin{equation}
\begin{array}{ll}
\vec{\lambda}_{(t)} = \vec{\lambda}_{(t-1)} + \gamma_t.\mathcal{I}(.)^{-1}.\mathop{\mathbb{E}}\Bigl[\nabla_{\vec{\lambda}} \log[g(\vec{\lambda})]\Bigr] \\
\vec{u}_{i_k = n,(t)}^{(k)} = \vec{u}_{i_k=n,(t-1)}^{(k)} + \gamma_t.\mathcal{I}(.)^{-1}.\mathop{\mathbb{E}}\Bigl[\nabla_{\vec{u}_{i_k=n}^{(k)}} H(\vec{u}_{i_k=n}^{(k)})\Bigr] \\
\vec{\beta}^{(k)}_{l, (t)} = \vec{\beta}^{(k)}_{l, (t-1)} + \gamma_t.\mathcal{I}(.)^{-1}.\mathop{\mathbb{E}}\Bigl[\nabla_{\vec{\beta}_l}^{(k)} F(\vec{\beta}_l^{(k)})\Bigr],
\end{array}
\label{sgd:eqn}
\end{equation}
} where the learning rate $\gamma_t$ in (\ref{sgd:eqn}) is given by:
\begin{equation*}
\gamma_t = \frac{1}{(\tau_{p} + t)^{\theta}}.
\end{equation*}
Note that $\theta$ is the forgetting rate and $\tau_p$ is the delay. The values for these are chosen such that $\sum_t \gamma_t^{2}$ is bounded but $\sum_t \gamma_t$ is unbounded. 
$\mathcal{I}(.)^{-1}$ in (\ref{sgd:eqn}) indicates the inversion of the Fisher information matrix (square matrix) in each iteration. There are two difficulties:
\newline
1) Computation of the Fisher information matrix is usually not trivial since it may not lend itself in a nice closed form.
\newline
2) Size of the Fisher information matrix in our data could possibly be in the tens of thousands or more. This impacts scalability, i.e., could result in very expensive computations that might also pose numerical stability issues leading to an intractable inverse computation.

For the first issue, we show that the P\'{o}lya-Gamma data augmentation facilitates easy computation of the Fisher information matrix. The detailed derivations of the partial Fisher information matrices as well as the gradients computations for each of the arguments, namely, $\vec{\lambda}$, $\vec{u}^{(1)}_{1} \ldots \vec{u}^{(K)}_{n_K}$ and $\vec{\beta}_l^{(1)} \cdots \vec{\beta}_L^{(K)}$ are sections \ref{supp_sec:fisher_matrix} and \ref{supp_sec:gradient}.


We have addressed the second issue by exploiting the problem structure which facilitates working with partial Fisher information matrix (hence called partial natural gradients). Partial natural gradients implies that we only work with diagonal blocks of the Fisher information matrix instead of the full matrix. Note that in our problem structure, the loss function is quadratic in each of the arguments ($\vec{\lambda}, \vec{u}^{(1)}_{1}, \allowbreak \ldots, \vec{u}^{(K)}_{n_K}, \vec{\beta}_l^{(1)}, \ldots, \vec{\beta}_L^{(K)}$). Due to the \emph{individually} quadratic nature of the loss functions, each diagonal block is a symmetric positive definite matrix of size $R \times R$ for $\vec{\lambda}$ and $\vec{u}^{(k)}_{n_k}$ or $(R+1) \times (R+1)$ for $\vec{\beta}_l^{(k)}$. Hence, the basic convergence guarantees for the full natural gradient learning extends to the partial set up as well \cite{bottou16a}. We note that computation of the partial Fisher information matrix is theoretically and numerically tractable as we are dealing with square matrices of size $R$ or ($R+1$), which is very small (value less than $10$) in our problem space.

For scalability over very large data sets, typical of Amazon data, we have implemented partial natural gradient learning in a stochastic setting. A concern here is that in each iteration, using a mini-batch, we are obtaining a noisy estimate of the partial Fisher information matrix. A noisy estimate of this matrix leads to a biased inverse since the mean of inverses is not equal to the inverse of the mean. Hence, theoretical guarantees regarding faster convergence for partial natural gradients in a stochastic setting as compared with stochastic gradients cannot be established. Another side-effect of this noisy estimate is that the matrix may be highly ill-conditioned, and hence may lead to numerical stability problems while computing its inverse. To circumvent this issue, we have experimented with the conditioning of this matrix (conditioning implies adding a scaled diagonal matrix) in either of the two ways. We can either use the prior of the corresponding latent variable to scale the identity matrix and add this to the partial Fisher information matrix (denoted as natural gradient 1). Or we can use the diagonal of the partial Fisher information matrix to scale the identity matrix and add this to the partial Fisher information matrix (denoted as natural gradient 2). 

So far, we have not come across any efforts that have applied partial natural gradient learning in a stochastic setting for inference in Bayesian CP tensor decomposition. In section \ref{sec:experiments}, we empirically show that in the semi-supervised setting, on the test data, scalar updates of vector parameters in Online-EM lead to very slow convergence and performance is sub-optimal w.r.t. ROC-AUC as compared with the partial natural gradient and stochastic gradient algorithms. We also empirically show that the partial natural gradient algorithm in a stochastic setting achieves faster convergence than stochastic gradient learning in detecting both abusive sellers and reviewers on Amazon data sets.





\section{Experimental Results}
\label{sec:experiments}
This section outlines our experiments to validate the efficacy of SENTINEL towards detecting entities promoting review abuse.\\
{\bf Dataset:} For all our experiments, we have taken a random sample of products from the Amazon reviews dataset between 2017 and 2018. Not all reviews are associated with the purchase of a product. As a result, considering only those reviews with an associated seller reduces the dataset size by half. From the review data, we construct two datasets and correspondingly two separate binary tensors - 1) {\em SELLER-TENSOR} captures the existing association of a reviewer $r$ giving a numeric rating $n$ at time $t$ for a product $p$ from a seller $s$ (5-mode tensor), and 2) {\em NO-SELLER-TENSOR} captures the existing association of a reviewer $r$ giving a numeric rating $n$ at time $t$ for a product $p$, without the seller information (4-mode tensor). The former is used to detect abusive sellers, while the latter is used to detect abusive reviewers. The modes of the two binary tensor are reviewer ID, product ID, seller ID (if included), numeric rating and time. Note that numeric rating corresponds to an integer between $1$ to $5$ and that time is converted to a week index. \vspace{0.05in} \\
SELLER-TENSOR consists of ~25K reviewers, ~475K products and ~70K sellers. NO-SELLER-TENSOR consists of ~90K reviewers and ~1.4M products. SELLER-TENSOR has 1.2M tuples, resulting in an extremely sparse tensor with a density of $1.6*10^{-8}$. \\
{\bf Benchmark Methods:} To the best of our knowledge, the proposed binary multi-target semi-supervised enhancement for a CP decomposition and its inference using partial natural gradients are novel to this paper. However, tensor decomposition in general has been applied to detect abusive entities in multi-modal data in the past. Hence, we benchmark the performance of our proposed approach with the following tensor based approaches:
\newline
{\em BPTF}: Bayesian Poisson Tensor Factorization \cite{schein15a} is a Bayesian CP tensor decomposition approach assuming a Poisson likelihood. The authors in \cite{schein15a} have implemented the unsupervised BPTF model using a batch algorithm, but it does not seamlessly lend itself to semi-supervised extensions. 
\newline
{\em M-Zoom}: The authors in \cite{shin16a} have proposed an unsupervised technique to identify dense blocks in tensors or dense sub-graphs in heterogeneous networks, which can also be applied to identify abusive sellers and reviewers.
\newline
{\em Unsupervised BNBCP}: Beta Negative-Binomial CP decomposition \cite{hu15a} is also a Bayesian CP tensor decomposition approach, assuming a Poisson likelihood. The BNBCP model is a fully conjugate model and inference is done using Variational Bayes (VB). To be able to scale for massive tensors, we have implemented an online VB version using Stochastic Variational Inference (SVI).
\newline
{\em Semi-supervised BNBCP}: Since the unsupervised BNBCP model can be easily extended to the semi-supervised setting, we implemented a semi-supervised version. Note that this is a single target and not a multi-target enhancement. 

We use precision, recall and AUC against a test set to measure the performance of the above approaches. \\
{\bf Experimental Validation:} Our experiments consist of the following empirical evaluations:
\newline
1)~~~Comparison between unsupervised and the proposed semi-supervised enhancements in detecting abusive sellers and reviewers. 
\newline
2)~~~Impact of different inference techniques on the proposed semi-supervised enhancement.
\newline
3)~~~Robustness of the proposed multi-target semi-supervised model to variations in the hyper-parameters.
\newline
4)~~~Stability of the proposed approach across many runs, owing to the non-convex nature of tensor decomposition. 
\newline
5)~~~Scalability of the proposed approach.
\newline
6)~~~Comparison with baseline models in terms of early detection of abusive reviewers and its impact on customers viewing the reviews.

Within SENTINEL, we have chosen the following values for the learning rate parameters $\tau_{p} = 256$ and $\theta = 0.61$. We have chosen the following values for the parameters of the \textit{Inverse Gamma} distributions: $a_c = 1$, $b_1 = 0.4$ and $b_2 = 3$. We have set the mini-batch size to $1024$ in all our simulations. 

\subsection{Detecting Abusive Sellers}
In this experiment, we take a random sample of sellers who have been flagged, via manual audits, for being guilty of review abuse. These are treated as positively labeled samples. In addition, we have included a random sample of sellers who are currently not flagged for any kind of abuse. These are treated as negatively labeled samples. Both these samples form the training dataset together. We have taken an additional set of around $1\%$ of sellers~\footnote{High confidence abusive sellers are hard to obtain due to business reasons, hence the small test set.} as test set to measure the performance of different techniques. The test set has a similar distribution as the training set and has been selected from beyond the training time period.
\begin{table}[t]
  \caption{Detecting abusive sellers: Comparison between benchmark methods. Note that the numbers are \textit{scaled}.} 
  \label{tab:unsem_s}
  \resizebox{\columnwidth}{!}{%
  \begin{tabular}{lccccl}
    \toprule
    & \textbf{Method} & \textbf{Precision} & \textbf{Recall} & \textbf{F1 Score} & \textbf{AUC}\\
    \midrule
    \multirow{3}{*}{Un-Supervised}
    & M-Zoom~\cite{shin16a} & \textbf{0.61} & 0.74 & \textbf{0.67} & -\\
    & BPTF~\cite{schein15a} & 0.53 & 0.79 & 0.63 & 0.74\\
    & BNBCP~\cite{hu15a} & 0.51 & \textbf{0.89} &  0.65 & \textbf{0.74}\\
    \midrule
    & BNBCP [\textit{SVI}] & 0.85 & 0.93 & 0.89 & 0.87\\
    \multirow{1}{*}{Semi-Supervised}
    & Logistic CP [\textit{Stochastic Gradient}] & \textbf{0.89} & 0.94 & \textbf{0.91} & \textbf{0.88}\\
    \multirow{1}{*}{(until convergence)}
    & Logistic CP [\textit{Natural Gradient 1}] & 0.85 & 0.93 & 0.89 & 0.88\\
    & Logistic CP [\textit{Natural Gradient 2}] & 0.84 & \textbf{0.94} & 0.89 & 0.88\\
    \midrule
    
    & Logistic CP [\textit{Stochastic Gradient}] & 0.81 & 0.74 & 0.77 & 0.78\\
    \multirow{1}{*}{Semi-Supervised}
    & Logistic CP [\textit{Natural Gradient 1}] & \textbf{0.83} & \textbf{0.89} & \textbf{0.86} & \textbf{0.87}\\
   \multirow{1}{*}{(stop at 200 iterations)}
    & Logistic CP [\textit{Natural Gradient 2}] & 0.78 & 0.81 & 0.79 & 0.85\\
  \bottomrule
\end{tabular}
}
\end{table}
Table \ref{tab:unsem_s} shows that all four flavors of the semi-supervised tensor decomposition have higher precision, recall and AUC as compared with the unsupervised techniques - indicating that leveraging behavioral patterns from current abusive sellers improves performance and fidelity in identifying new abusive sellers. Given the single type of seller abuse (aka single target); both semi-supervised BNBCP and SENTINEL models have comparable AUC performance. With early stopping of the semi-supervised models ($200$ iterations), we observe that inference based on natural gradient outperforms stochastic gradient learning, indicating empirically faster convergence of the former.

Note that M-Zoom does not produce the scores for each suspicious entity and hence the AUC is not calculated. Due to this lack of scores per entity, we cannot rank the suspicious entities to be able to apply different levels of enforcement actions against them. Among the unsupervised methods, BNBCP has the best AUC, which is around $16$\% lower than the best semi-supervised approach. The best performing unsupervised method in terms of precision is M-Zoom, which is around $46$\% lower than the best semi-supervised approach. 

Table \ref{tab:selr} shows evidence from the review data indicating suspicious behavior for a representative set of sellers that was predicted by SENTINEL to be abusive. Density indicates the ratio of bipartite connections (reviews) observed in the data to the maximum number of bipartite connections between products from this seller and suspicious reviewers. These reviewers are most likely abusive since they have written a review for practically all products by the seller.
\begin{table}[t]
  \caption{Abusive sellers: evidence from review data.}
  \label{tab:selr}
  \resizebox{\columnwidth}{!}{%
  \begin{tabular}{lccccl}
    \toprule
    & \textbf{Suspicious Seller} & \textbf{\# Suspicious Products} & \textbf{\# Suspicious Reviewers} & \textbf{\# Reviews} & \textbf{Density}\\
    & A & 75 & 25 &  1802 & 0.96\\
    & B & 16 & 9 & 132 & 0.91\\
    & C & 7 & 19 & 114 & 0.86\\
  \bottomrule
\end{tabular}
}
\end{table}
Since paid reviewer abuse mostly manifests off-Amazon (e.g., funds transferred through bank), evidence gathered from review data is indicative but not legally binding to qualify the sellers as fraudulent. Moreover, since enforcement actions on false positives would impact revenue negatively, the thresholds for enforcement are quite stringent. Majority of the sellers are warned based on the model's recommendation. 


Section \ref{supp_sec:CI} shows confidence intervals of the algorithms.

\subsection{Detecting Abusive Reviewers}
In this experiment, we have taken a random sample of reviewers, who have been identified to be guilty of review abuse. Among these, roughly $60$\% belong to paid reviewer abuse category and the remainder belong to the compromised account abuse category. Recall that the compromised account abuse occurs when a good customer's account is taken over by an abusive reviewer to post fake reviews. SENTINEL supports multi-target labels in a semi-supervised setting. Hence, we treat the two forms of abuse as two separate targets in our model. The BNBCP semi-supervised model, on the other hand, does not support multiple targets. Hence there is no differentiation between these two forms of abuse in the BNBCP semi-supervised model. A random sample of about $80$\% of this data forms the training set and the remaining $20$\% forms the test set. 
\begin{table}[t]
  \caption{Detecting abusive reviewers: Comparison between benchmark methods. Note that the numbers are \textit{scaled}.}
  \label{tab:unsem_r}
  \resizebox{\columnwidth}{!}{%
  \begin{tabular}{cccccl}
    \toprule
    & \textbf{Method} & \textbf{Precision} & \textbf{Recall} & \textbf{F1 Score} & \textbf{AUC}\\
    \midrule
    \multirow{3}{*}{Unsupervised}
    & M-Zoom~\cite{shin16a} & 0.53 & \textbf{0.70} & \textbf{0.60} & -\\
    & BPTF~\cite{schein15a} & 0.42 & 0.54 & 0.47 & 0.51\\
    & BNBCP~\cite{hu15a} & \textbf{0.59} & 0.46 &  0.52 & \textbf{0.61}\\
    \midrule
    & BNBCP [\textit{SVI}] & 0.60 & 0.83 & 0.69 & 0.79\\
    \multirow{1}{*}{Semi-Supervised}
    & Logistic CP [\textit{Stochastic Gradient}] & 0.60 & 0.90 & 0.72 &  0.85\\
   \multirow{1}{*}{(until convergence)}
    & Logistic CP [\textit{Natural Gradient 1}] & \textbf{0.62} & 0.91 & \textbf{0.74} & \textbf{0.85}\\
    & Logistic CP [\textit{Natural Gradient 2}] & 0.59 & \textbf{0.91} & 0.72 & 0.85\\
    \midrule
    & Logistic CP [\textit{Stochastic Gradient}] & 0.56 & 0.79 & 0.66 &  0.72\\
    \multirow{1}{*}{Semi-Supervised}
    & Logistic CP [\textit{Natural Gradient 1}] & \textbf{0.59} & \textbf{0.87} & \textbf{0.71} & \textbf{0.83}\\
   \multirow{1}{*}{(stop at 600 iterations)}
    & Logistic CP [\textit{Natural Gradient 2}] & 0.58 & 0.86 & 0.69 & 0.82\\
  \bottomrule
\end{tabular}
}
\end{table}
Table \ref{tab:unsem_r} shows that all the four variants of our semi-supervised tensor decomposition have higher precision, recall, and AUC as compared with the unsupervised techniques. Recall and AUC have shown significant increase between the unsupervised and semi-supervised methods, as compared to the precision. With early stopping ($600$ iterations), we see that the natural gradient outperforms stochastic gradient learning, indicating empirically faster convergence of the former. At full convergence, multi-target logistic CP semi-supervised model exhibits better performance as compared with the semi-supervised model that is not designed to differentiate between different forms of abuse (i.e., BNBCP model). To test the hypothesis that jointly learning separate models for each abuse type is better than learning a single model for all abuse types, we experimented with using a single target which is the union of the two abuse types in SENTINEL. Results indicated that we see an almost $3$\% gain in AUC in predicting one of the targets in the multi-target scenario, hence confirming our hypothesis.


\subsection{Stochastic Partial Natural Gradients versus Baseline Learning Methods} 
\begin{figure}[H]
\centering
\begin{tabular}{cc}
\noindent \begin{minipage}[b]{0.5\hsize}
\centering
\includegraphics[width=1.0\textwidth]{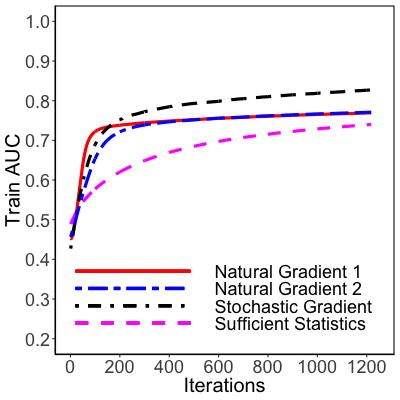}\\
{(a) Train data: sellers.}
\end{minipage}
\begin{minipage}[b]{0.5\hsize}
\centering
\includegraphics[width=1.0\textwidth]{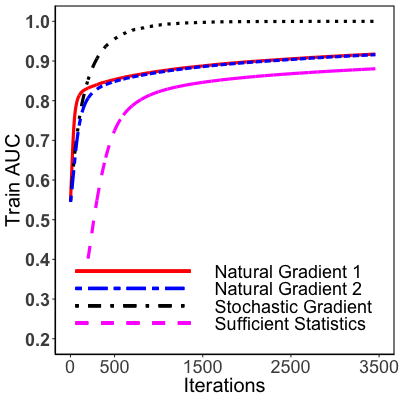}\\
{(b)  Train data: reviewers.}
\end{minipage}\\
\noindent \begin{minipage}[b]{0.5\hsize}
\centering
\includegraphics[width=1.0\textwidth]{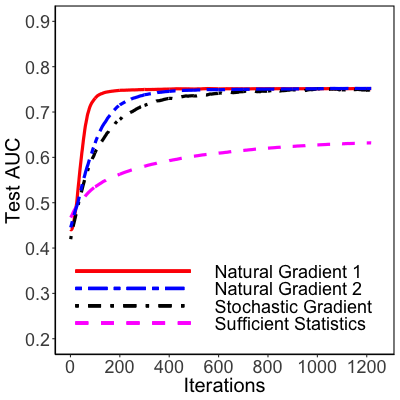}\\
{(c) Test data: sellers.}
\end{minipage}
\begin{minipage}[b]{0.5\hsize}
\centering
\includegraphics[width=1.0\textwidth]{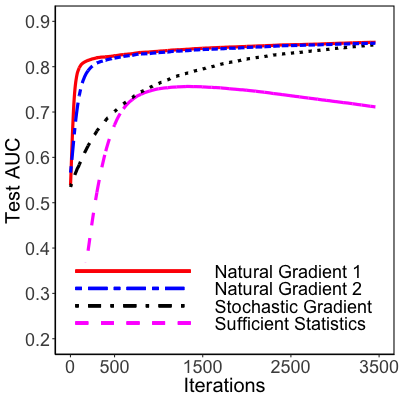}\\
{(d)  Test data: reviewers.}
\end{minipage}
\end{tabular}
\caption{Efficiency of partial natural gradient learning in identifying abusive sellers and reviewers.}
\label{ngdSgd:fig}
\end{figure}

We apply multi-target semi-supervised Logistic CP tensor decomposition approach on Amazon review data (SELLER-TENSOR for sellers and NO-SELLER-TENSOR for reviewers) to compare the performance of two baseline learning methods, namely, sufficient statistics (Online-EM) and stochastic gradient with the proposed stochastic partial natural gradient learning. As mentioned in Section \ref{sec:Mstep}, we have two flavors of stochastic partial natural gradient learning, differing only by the conditioning applied to the partial Fisher information matrix. Figure~\ref{ngdSgd:fig} shows the ROC-AUC plot for detecting abusive sellers and abusive reviewers versus the iteration number. Solid red plot corresponds to the stochastic partial natural gradient learning type 1, blue (dash-dot) plot corresponds to stochastic partial natural gradient learning type 2, black dashed plot corresponds to stochastic gradient learning and magenta (dash) plot corresponds to online EM with sufficient statistics. 

Stochastic gradient learning has a tendency to over-train since it is unable to shrink some of the elements of $\vec{\lambda}$ towards zero as the tensor rank is less than $R$. Stochastic partial natural gradient learning (both flavors) does not suffer from significant over-training since it is able to shrink 60\% of the values of $\lambda_r$ towards zero within the first one thousand iterations. This leads to similar AUC on train and test data sets for detecting both abusive sellers and abusive reviewers as compared with the other two baselines. Stochastic partial natural gradient learning type 2 has a slightly slower learning rate than type 1. Online-EM with sufficient statistics shows poorer performance on test data (for both reviewers and sellers) when compared with stochastic gradient or partial natural gradient learning.
\subsection{Sensitivity to Model Hyper-parameter}
\begin{figure}[H]
\centering
\begin{tabular}{cc}
\hspace{-0.3cm} 
\begin{minipage}[b]{0.5\hsize}
\centering
\includegraphics[scale=0.3]{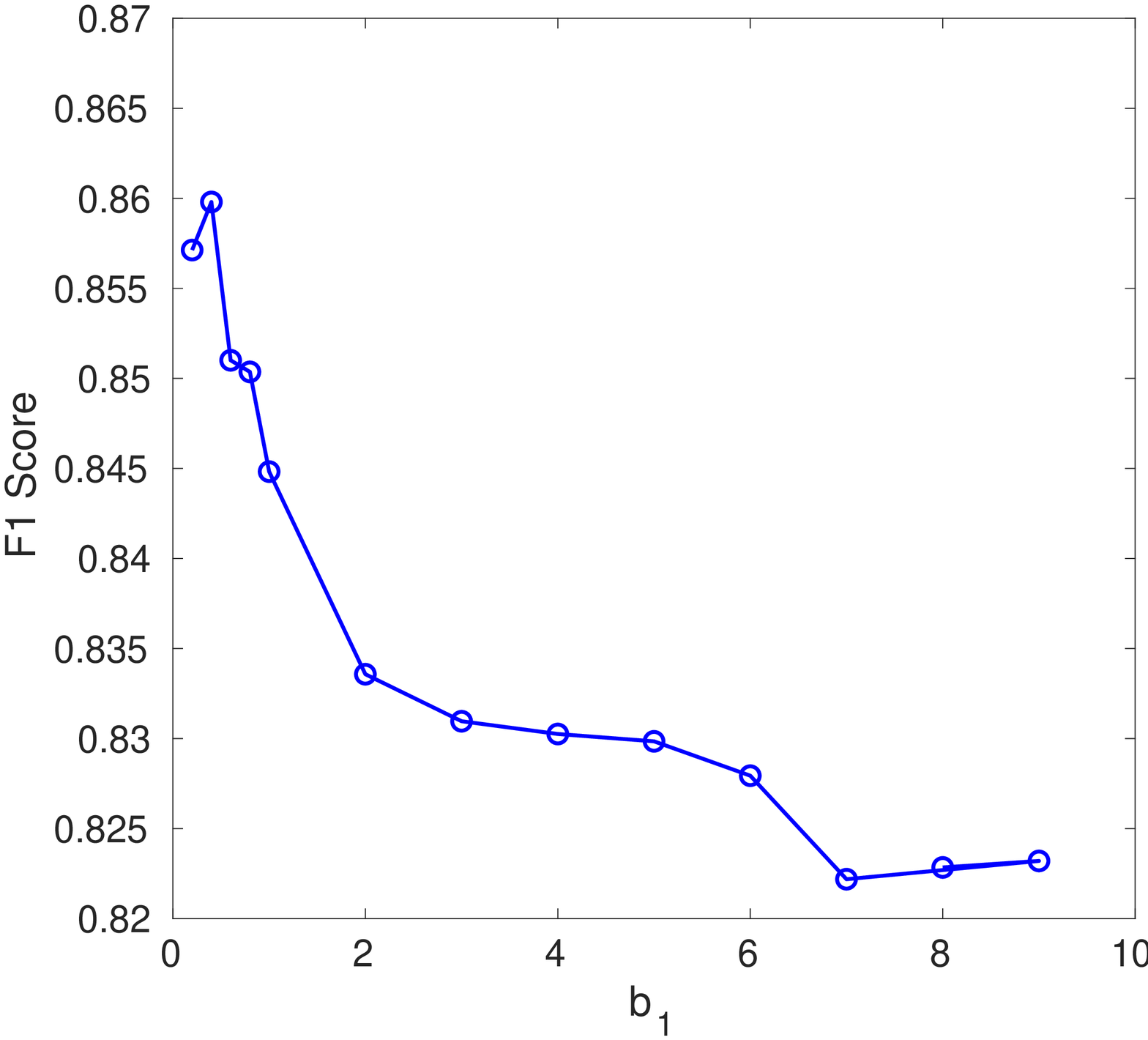}\\
{(a) F1 Score versus $b_1$ with $b_2$ Fixed at 9.}
\end{minipage}
\begin{minipage}[b]{0.5\hsize}
\centering
\includegraphics[scale=0.3]{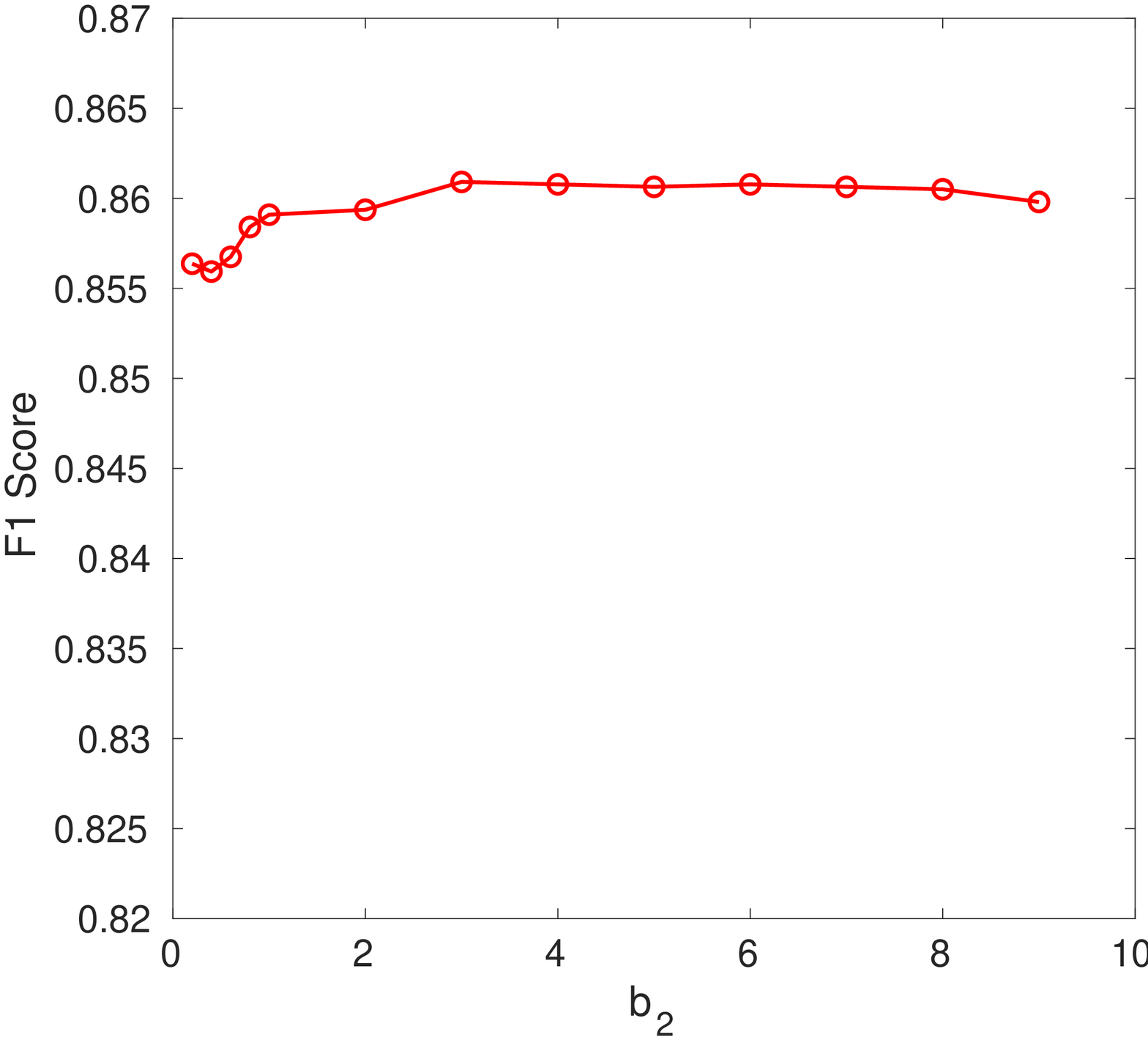}\\
{(b)  F1 Score versus $b_2$ with $b_1$ Fixed at 0.4.}
\end{minipage}
\end{tabular}
\caption{Impact of varying hyper-parameters on F1 Score.}
\label{hyperP:fig}
\end{figure}

Impact of the scale parameter of the two Inverse-Gamma distributions (i.e., model hyper-parameters), namely, $b_1$ and $b_2$ on F1 score in a semi-supervised setting is shown in Figure~\ref{hyperP:fig} ($a$) and ($b$). $b_1$ ($b_2$) is the scale parameter of the Inverse-Gamma distribution that controls the variance of the Gaussian prior for the factors that belong to the mode(s) associated with (without) multi-target data. $b_2$ is also the scale parameter for the Inverse-Gamma distribution that controls the variance of the Gaussian prior for the Logistic regression coefficients corresponding to each task $l$. $b_2$ is set such that it offers very little regularization, i.e., set to a high value. However, $b_1$ has to be chosen more carefully such that it provides the right amount of regularization in the semi-supervised setting. 
In Figure~\ref{hyperP:fig} ($a$), we fix $b_2$ to a arbitrarily high value and vary $b_1$ from $0.2$ until $9$. We notice that a value of $b_1 = 0.4$ produces the highest F1 score. In Figure~\ref{hyperP:fig} ($b$), we fix $b_1 = 0.4$ and vary $b_2$ from $0.2$ until $9$. We notice that the impact on F1 score of varying $b_2$ is minimal as compared with varying $b_1$ and that the best F1 score is achieved at $b_2 = 3.0$. Hence all our experiments are done with $b_1 = 0.4$ and $b_2 = 3.0$. 

\subsection{Robustness of Various Techniques}
\begin{figure}[H]
\centering
\begin{tabular}{cc}
\hspace{-0.3cm} 
\begin{minipage}[b]{0.5\hsize}
\centering
\includegraphics[scale=0.3]{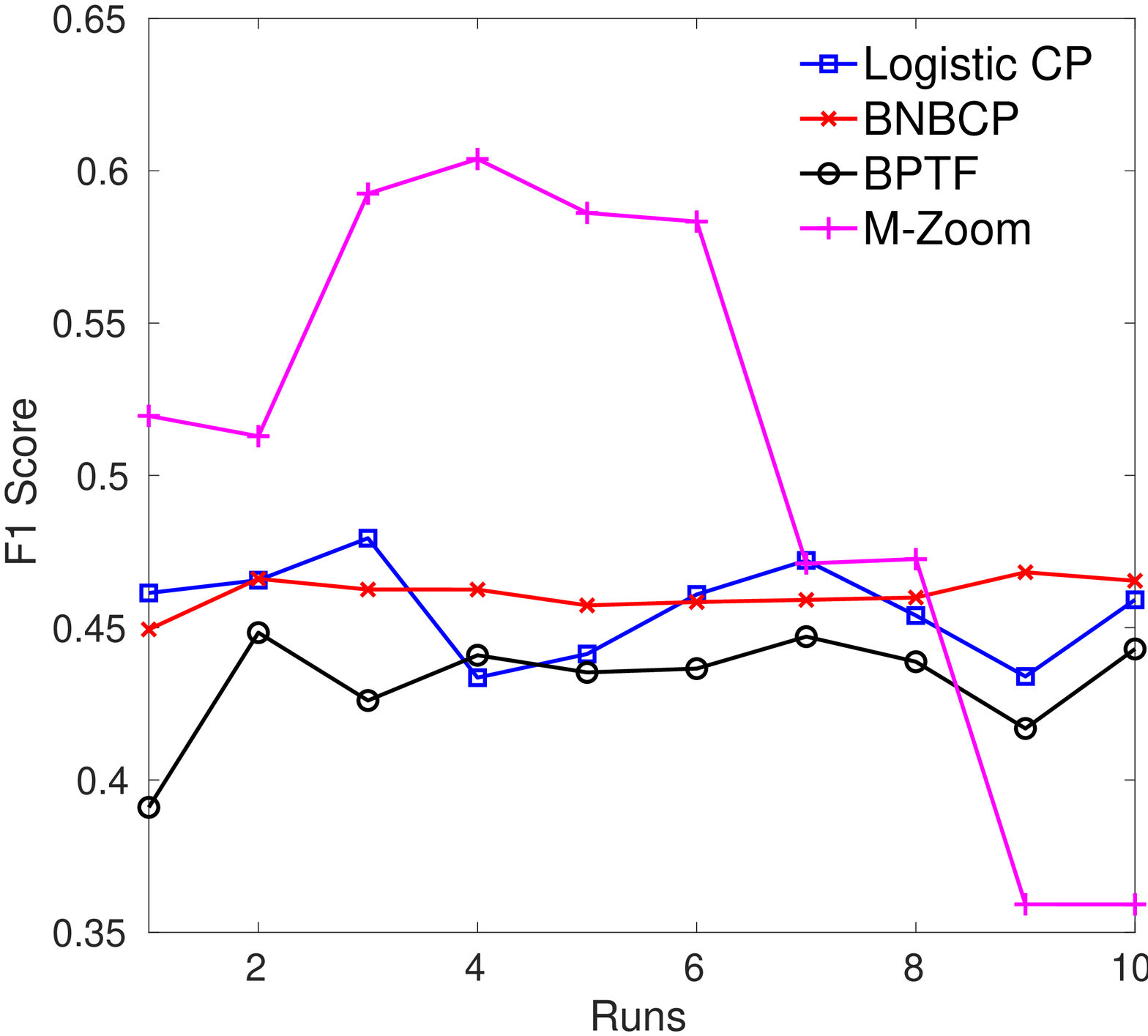}\\
{(a) Robustness: F1 Scores across Runs.}
\end{minipage}
\begin{minipage}[b]{0.5\hsize}
\centering
\includegraphics[scale=0.3]{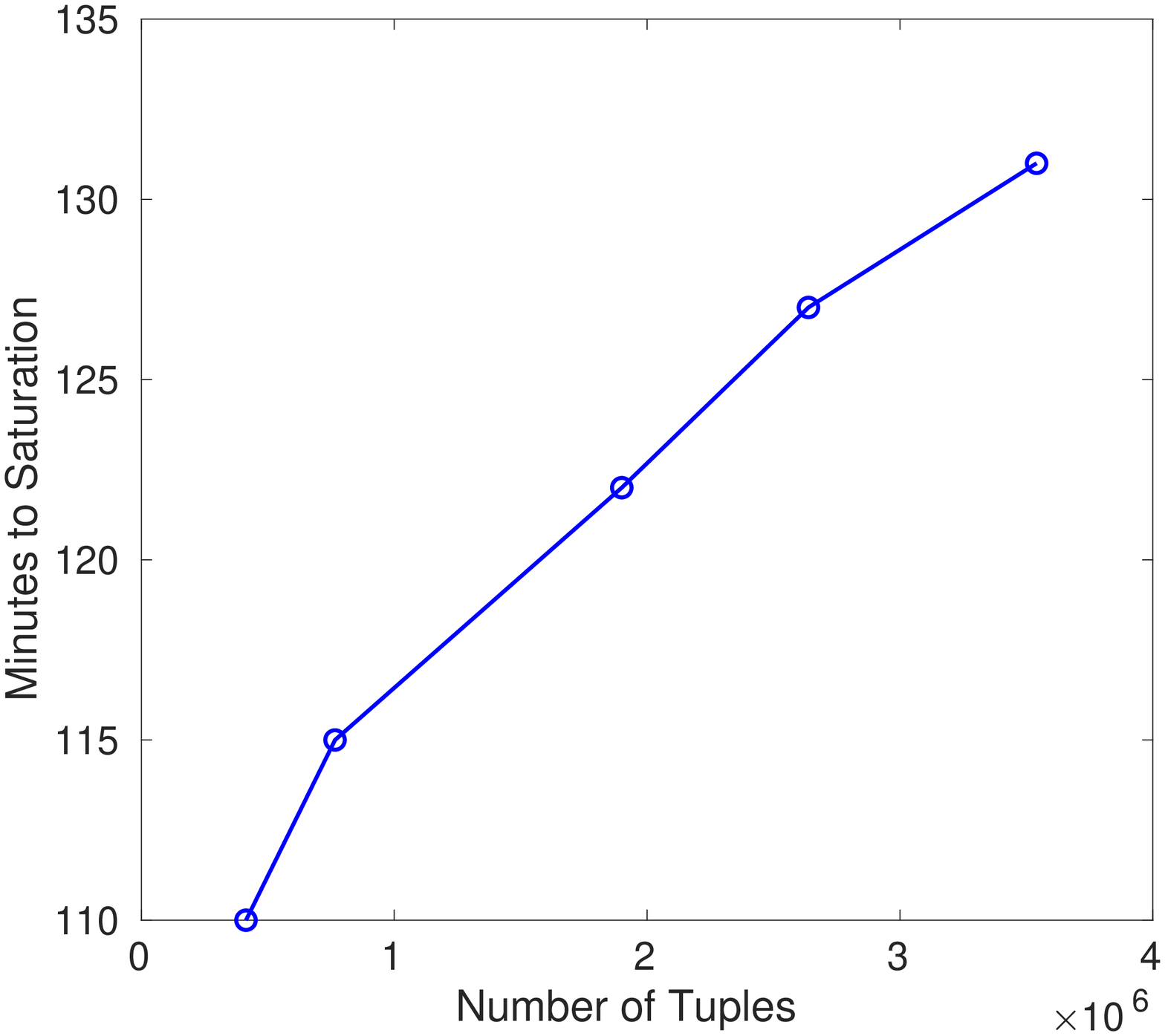}\\
{(b)  Scalability: Size of Tensor vs Convergence Time.}
\end{minipage}
\end{tabular}
\caption{Robustness and Scalability.}
\label{variab:fig}
\end{figure}

Due to the non-convex nature of tensor decomposition, Figure~\ref{variab:fig} ($a$) shows the variations in the F1 scores across ten different runs for each of the techniques, namely, Logistic CP with stochastic partial natural gradient inference, BNBCP, BPTF and M-Zoom (unsupervised setting) for detecting abusive reviewers. 

BNBCP has the least variations across different runs, followed by the Logistic CP with stochastic partial natural gradient based inference.
M-Zoom (in magenta) requires the number of blocks to be specified as input. Given the number of blocks, it produces identical sub-tensors across different runs. Hence, we have measured the F1 score performance across different number of blocks, monotonically varying them from $5$ to $14$. Below $5$, the F1 score was much lower, hence it is omitted. We see that with $8$ blocks as input, the F1 score is at the highest and it beats the other three methods. However, unlike the other three methods, M-Zoom does not produce a score for each suspicious reviewer. That is, we cannot rank the reviewers according to their suspiciousness (that shows that M-Zoom lacks a very important factor for taking enforcement actions). 

\subsection{Scalability of the proposed approach}
We ran the logistic CP with partial natural gradient on the NO-SELLER-TENSOR dataset with varying number of tuples. The hyper-parameters were fixed at the same setting mentioned above. Figure~\ref{variab:fig} ($b$) shows the scalability of the proposed method across five different datasets of varying sizes. We plot the time to converge (iteration number at which learning is saturated) in minutes versus the size of the tensor (i.e., the number of tuples) in log scale. We notice an almost linear increase in computational time to converge as the number of tuples increases.
Our algorithm is also easily parallellizable because in a given iteration, each tuple in the mini-batch can be processed independently. This, we expect should help us in achieving much faster convergence than what we are currently seeing. We are in the process of implementing a parallel version of our algorithm. 

\subsection{Impact of Early Detection of Abusive Reviewers}
\begin{figure}[H]
\centering
\begin{tabular}{cc}
\hspace{-0.3cm} 
\begin{minipage}[b]{0.5\hsize}
\centering
\includegraphics[scale=0.3]{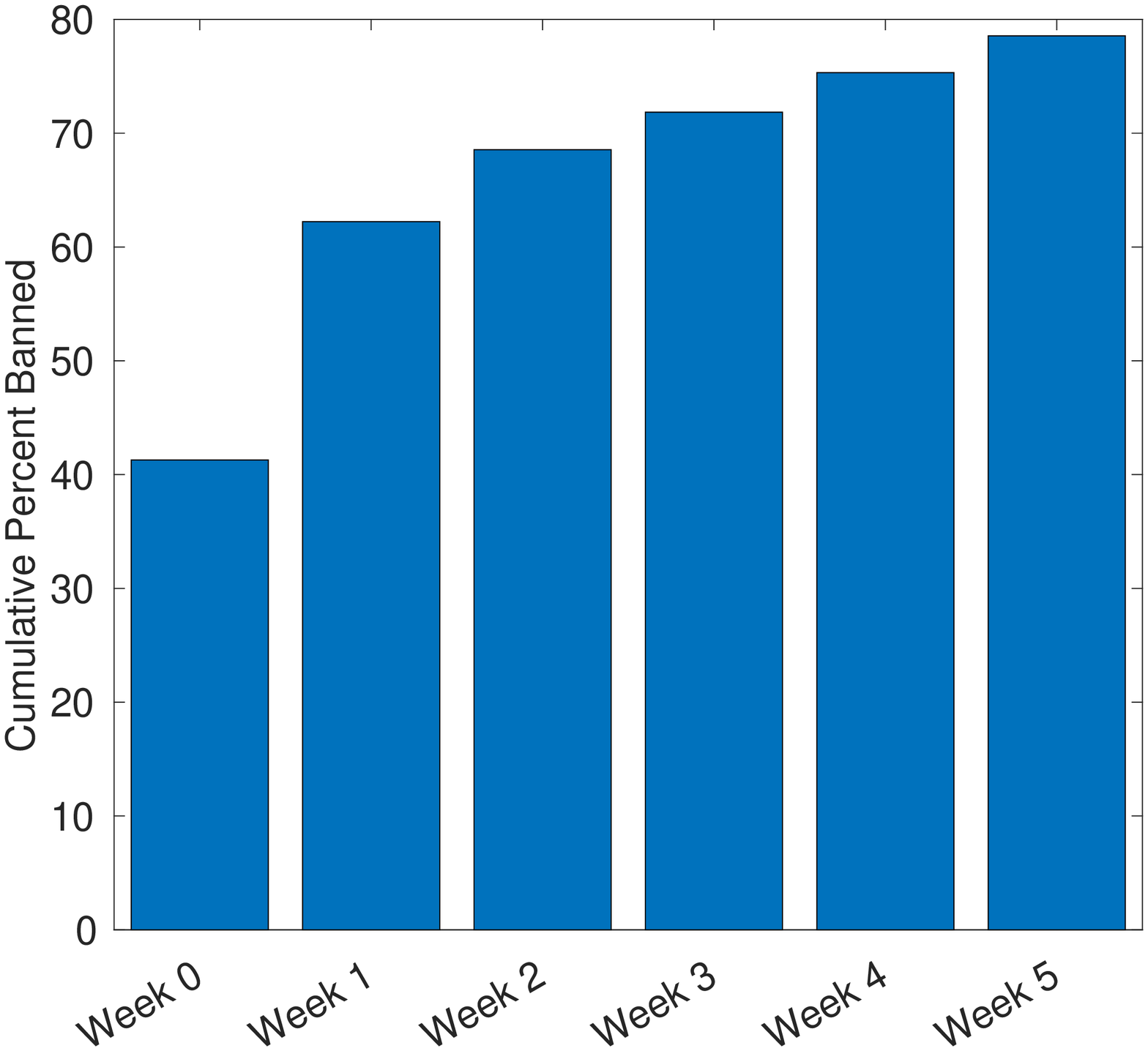}\\
{(a) Cumulative Percentage of Reviewers Banned versus Time.}
\end{minipage}
\begin{minipage}[b]{0.5\hsize}
\centering
\includegraphics[scale=0.3]{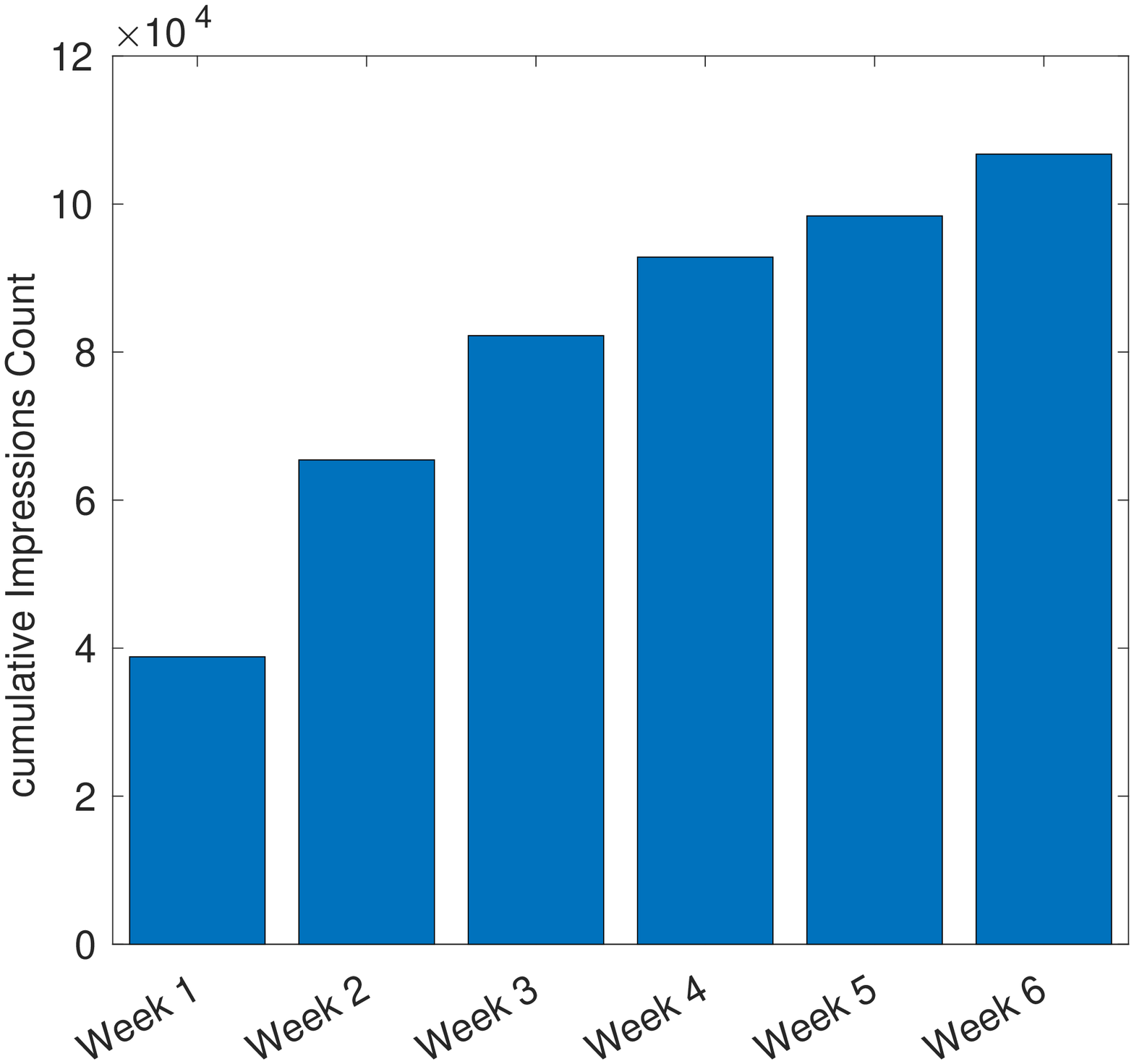}\\
{(b)  Cumulative Impressions Count versus Time.}
\end{minipage}
\end{tabular}
\caption{Impact of Early Detection.}
\label{fracBan:fig}
\end{figure}

In this section, we present a retrospective analysis of the impact of our model as compared with the existing system. \\
\textbf{Early detection of abusive reviewers}: Figure~\ref{fracBan:fig} ($a$) shows the fraction of suspicious reviewers predicted by SENTINEL, that are later banned by existing models. The model is executed during '$Week_0$' and $~40$\% of the identified suspicious reviewers overlap with those banned by production models. This overlap increases as we move forward by a week at a time, until five weeks later (indicated by \textit{Week 5}) when the overlap is roughly $80$\%. This indicates that roughly $40$\% of the reviewers that are detected by our model at \textit{Week 0} are detected across next $5$ weeks by the production system. The impact of not banning the abusive reviewers at \textit{Week 0} is captured by the number of impressions created by the reviews from those abusive reviewers until \textit{Week 5}.\\
\textbf{Impact in terms of affected impression counts}: Impressions implies the number of customers who read the reviews (i.e., fake reviews) from those abusive reviewers until they were banned. Note that banning a reviewer results in removal of all his/her reviews. Hence larger the impression count implies more people have read those reviews which means bigger impact from not banning these abusive reviewers early enough. Figure~\ref{fracBan:fig} ($b$) shows the cumulative impact in terms of the number of impressions accumulated over consecutive weeks, from the time they were detected by our model until their banning by existing production model. These impressions on fake reviews quantify the potential impact of our model, once deployed in production. 

\subsection{Experiments on public data} 
To allow others to reproduce our results, we applied SENTINEL to the public Amazon Customer Reviews dataset\footnote{The dataset is available at {{{\url{http://jmcauley.ucsd.edu/data/amazon/}}.}}}~\cite{amazon-dataset}. We selected the product categories \textit{Clothing, Shoes, and Jewelry}, \textit{Home and Kitchen} and \textit{Sports and Outdoors} for this test. Note that this data does not have seller information. Also since this data is from May $1996$ to July $2014$, we do not have any labelled data, i.e., known abusive reviewers corresponding to that period. To test SENTINEL, we used random $80$\% of the suspicious reviewers from M-Zoom to seed SENTINEL as well as the semi-supervised version of BNBCP. The remaining $20$\% of the suspicious reviewers were considered as the test set. On the test set, we obtain a $5.6$\% increase in AUC with SENTINEL as compared with the semi-supervised BNBCP model and $5.9$\% increase in AUC compared with the best unsupervised model, i.e., BPTF. Since we are considering M-Zoom's output as the ground truth, we acknowledge that this is not an ideal setting to validate the performance of the proposed approach. Having said that, this setup at least allows us to loosely compare various methods.


\section{Conclusion}
\label{sec:conclusions}
We formulated the problem of identifying abusive entities (i.e., sellers and reviewers) as a tensor decomposition problem and proposed semi-supervised enhancements by incorporating binary multi-target information for a subset of entities, i.e., known abusive sellers and/or reviewers. Our results demonstrated that the proposed approach (titled SENTINEL) beats the state-of-the-art baselines in detecting abusive entities.

We have shown, in the supplementary manuscript, that P\'{o}lya-Gamma formulation simplifies calculation of the partial Fisher information matrix, and hence proposed a scalable stochastic partial natural gradient learning for inference of all the latent variables of the semi-supervised model. We have empirically shown that our inference using stochastic partial natural gradient learning achieves faster convergence than online EM using sufficient statistics and stochastic gradient learning. \\
{\bf Future Work:} Given the equivalence between tensor decomposition and convolutional rectifier networks \cite{cohena}, we would like to compare the performance of the latter (hierarchical Tucker) with our probabilistic CP decomposition model. We also want to investigate the feasibility of applying \textit{Graph Convolutional Networks} to our multi-modal data (with data being either reviewers and products and their relationships w.r.t. to rating as well as time of rating OR reviewers, products and sellers and their relationships) to detect abusive entities.

\section*{Acknowledgment}
This work was started at Amazon under the guidance of Srinivasan H. Sengamedu who suggested application of semi-supervised tensor decomposition based techniques towards review abuse detection. We would also like to thank Purushottam Kar for sharing helpful insights related to the biased estimates of the partial Fisher information matrices in our stochastic algorithm.

\bibliographystyle{amsalpha}
\bibliography{tensorcompletion}

\clearpage 
%
\appendix

\section{APPENDIX}

\subsection{Determine $\vec{\lambda}$}\label{sec:detLam}
$\vec{\lambda}$ has a Gaussian prior whose variance is determined via Multiplicative Gamma Process \cite{bhattacharya11a, durante16a} aka \textit{Adaptive Dimensionality Reduction} technique that induces sparsity and automatically deletes redundant parameters. The Multiplicative Gamma Process consists of a multiplicative Gamma prior on the precision of a Gaussian distribution that induces a multiplicative Inverse-Gamma prior on its variance parameter. Since its performance is sensitive to the hyper-parameter settings of the Inverse-Gamma prior we follow certain strategies, shown in \cite{durante16a}, to set their values.
The generative model for $\lambda_r$ for $r$ in $[1, R]$ and the Multiplicative Gamma Process are:
\begin{eqnarray*}
\delta_l \sim \text{Inv-Gamma}(a_l, 1) \\
\lambda_r \sim \mathcal{N}(0, \tau_r),
\end{eqnarray*}
where:
\begin{equation} \label{tauEq:eqn}
\tau_r = \prod_{l=1}^{r} \delta_l.
\end{equation}
The idea is that for increasing \textit{r}, $\tau_r$ should be decreasing in a probabilistic sense. In other words, the following stochastic order should be maintained with high probability, i.e.,
\begin{equation*}
\tau_1 \geq \tau_2 \geq \cdots \geq \tau_R.
\end{equation*}
To guarantee such a stochastic order with a high probability, as suggested in \cite{durante16a}, we need to set $a_1 > 0$ and $a_r > a_{r-1}$ and non-decreasing for all $r > 1$. Hence, we choose the hyper-parameters values as follows:
\begin{eqnarray*}
a_1 = 1 \\
a_r = a_1 + (r - 1).\frac{1}{R}.
\end{eqnarray*}
The update for $\delta_r$ in iteration \textit{t} is given by (see  \cite{rai14a} for details):
\begin{equation}
\delta_r = \frac{1 + \sum_{h = r}^{R} \frac{\lambda_h^{2}}{2} \prod_{l = 1, l \neq r}^{h} \frac{1}{\delta_l}}{0.5(R - r + 1) + a_r + 1}. \label{deltaEq:eqn}
\end{equation}
From (\ref{tauEq:eqn}), we can calculate the $\tau_r$s in iteration \textit{t} for $r \in [1,R]$. Denote $\vec{\tau}$ as the vector consisting of $\tau_r$ for $r \in [1,R]$.

Let $\lambda_r$ for $r \in [1,R]$ denote an element of $\vec{\lambda}$. We introduce auxiliary variables (P\'{o}lya-Gamma distributed variables \cite{polson, pillow1}), denoted by $\omega_i$ for each element $i$ of the input tensor data, via data augmentation technique.

Consider a mini-batch defined at iteration $t$ as $I_t$. Define for each $i \in I_t$ and $\phi_i = \vec{\lambda}^{T}A_i$ where: $A_i$ denotes a vector consisting of elements $A_i^{r} = \prod_{k = 1}^{K}u_{i_k,r}^{(k)}$ for $r \in [1, R]$. Let $A$ be the matrix whose rows are $A_i$ for $i \in I_t$. Let $\hat{\omega}_i$ for $i \in I_t$ be the expected value of the auxiliary variable $\omega_i$ corresponding to the $i^{th}$ element of the input tensor data. The expected value of $\omega_i$ has a closed-form solution given by:
\begin{equation*} E[\omega_i] = \hat{\omega}_i = \frac{\tanh(\frac{\phi_{i}}{2})}{2\phi_{i}}. \end{equation*}
The update for $\vec{\lambda}$ in iteration \textit{t}, with the current mini-batch $I_t$, is obtained by maximizing the natural logarithm of the posterior distribution of $\vec{\lambda}$ given by:
{\small
\begin{equation}
\log[g(\vec{\lambda})] := \max_{\vec{\lambda}} \Bigl[ \Bigl[\sum_{i \in I_t} \kappa_i\phi_i - \frac{\omega_i\phi_i^{2}}{2}\Bigr] - \Bigl[\frac{(\vec{\lambda} \oslash \sqrt{\vec{\tau}})^\top(\vec{\lambda} \oslash \sqrt{\vec{\tau}})}{2}\Bigr] \Bigr],
\label{fMax:eqn}
\end{equation}
}where $\kappa_{i} = y_i - \frac{1}{2}$ and $y_i \in \{0, 1\}$. And the operator $\oslash$ represents element-wise division between the two vectors $\vec{\lambda}$ and $\vec{\tau}$.
Note that (\ref{fMax:eqn}) is a quadratic equation in $\phi_i$ i.e., $\vec{\lambda}$, and hence has a \emph{closed-form} update.

\subsection{Determine the coefficients $\vec{\beta}_{l}^{(k)}$ for $l \in [1, L]$}\label{sec:detBeta}
The generative model for $\beta_{l,r}^{(k)}$ for $r$ in $[0, R]$, $l$ in $[1, L]$ is:
\begin{eqnarray*}
\rho_{l,r}^{(k)} \sim \mbox{Inv-Gamma}(a_c, b_2), \\
\beta_{l,r}^{(k)} \sim \mathcal{N}(0, {\rho_{l,r}^{(k)}}^{2}).
\end{eqnarray*}
For mode $k$, we drop the notation $(k)$ for all variables. Define for each $i \in M$ where $M$ denotes the number of elements in mode $k$ that have binary side information corresponding to one or more targets. The logistic function i.e., the likelihood $\mathcal{L}_{l,m}$ corresponding to element $m$ with label $z_{l,m}$ is given by:
\begin{equation*} \mathcal{L}_{l,m} = \frac{1}{1 + exp[-z_{l,m}{\vec{\beta}_{l}^{\top}\tilde{\vec{u}}_{m}}]}, \end{equation*}
where $\tilde{\vec{u}}_m$ denotes $\vec{u}_m$ prepended with $1$ to account for the bias.
With introduction of P\'{o}lya-Gamma variables \cite{polson} denoted by $\vec{\nu}_{l}$ for task $l$, the likelihood with the data augmentation becomes:
\begin{equation*} \mathcal{L}_{l,m} = exp(z_{l,m}.\frac{\psi_{l,m}}{2} - \hat{\nu}_{l,m}.\frac{\psi_{l,m}^{2}}{2}), \end{equation*}
where:
\begin{equation}
\begin{split}
\psi_{l,m} = \vec{\beta}_{l}^{\top}(\tilde{\vec{u}}_{m}) \\
\mathbb{E}[\nu_{l,m}] = \hat{\nu}_{l,m} = \frac{tanh(\frac{\psi_{l,m}}{2})}{2\psi_{l,m}}
\label{fNuPG:eqn}
\end{split}
\end{equation}
The update for $\vec{\beta}_l$ in iteration \textit{t} is obtained by maximizing the natural logarithm of the posterior distribution of $\vec{\beta}_l$ given by:
\begin{equation}
\begin{split}
F(\vec{\beta}_l):=\max_{\vec{\beta}_l}\Bigl[
\sum_{m = 1}^{M} [z_{l,m}\frac{\psi_{l,m}}{2} - \nu_{l,m}\frac{{\psi_{l,m}^{2}}}{2}] \\
- \frac{(\vec{\beta}_l \oslash \vec{\rho}_l)^\top (\vec{\beta}_l \oslash \vec{\rho}_l)}{2} - \sum_{j \neq l} Q_{l,j}[\vec{\beta}_{l}^{\top}Sign(\vec{\beta}_{j})]
\Bigr].
\label{fBeta:eqn}
\end{split}
\end{equation}
Equation (\ref{fBeta:eqn}) is a quadratic equation in $\vec{\beta}_l$ and hence has a \emph{closed-form} update. And the operator $\oslash$ represents element-wise division between the two vectors $\vec{\beta}_l$ and $\vec{\rho}_l$.

Subsequently, the update for $\rho_{l,r}^{2}$ at time step $t$ is given by:
\begin{equation}\label{eq:update_rho}
{\rho_{l,r}^{2}}_{(t)} = \frac{{\beta_{l,r}^{2}}_{(t-1)}}{2a_c + 3} + \frac{2b_1}{2a_c + 3}.
\end{equation}

\subsection{Determine the factors $\vec{u}_{i_k,r}^{(k)}$ for mode $k$}\label{sec:detUfactor}
The generative model for $u_{i_k,r}^{(k)}$ is:
\begin{equation*}
\begin{array}{l}
u_{i_k,r}^{(k)} \sim \mathcal{N}(0, \mu_{i_k, r}^{2})\\
\mu_{i_k,r}^{2} \sim \text{Inv-Gamma}(a_c, b_1)\qquad  \text{with target information} \\
\mu_{i_k,r}^{2} \sim \text{Inv-Gamma}(a_c, b_2) \qquad   \text{without target information}.
\end{array}
\end{equation*}
The Inverse-Gamma hyper-prior on the variance parameter of the Gaussian prior for $u_{i_k,r}^{(k)}$ provides \textit{adaptive L2-Regularization}. The Inverse-Gamma parameters are set so that greater amount of regularization is provided for mode $k$ that has target information.
Denote $\vec{u}_{k,i_k}$ as the $R$ dimension vector consisting of factors $u_{i_k,r}^{(k)}$ corresponding to element $i_k$ in mode $k$ and $r \in [1, R]$. Define for each $i \in I_t$, $i_k = n$, and $\phi_i = (\vec{u}_{i_k = n}^{(k)})^{T}\vec{C}_{i_k=n}$, where
each element of $\vec{C}_{i_k=n}$ is $C_{i_k=n,r} = \lambda_r.\prod_{k^{'} \neq k}^{K} u_{i_{k^{'}},r}^{(k^{'})}$ for $r \in [1, R]$. Let $\boldsymbol{C}^{(k)}$ be the matrix whose rows are $\vec{C}_{i_k}$.

\textbf{Mode $k$ without target information}:
For mode $k$, we drop the notation $(k)$ for all variables. The update for $\vec{u}_{i_k=n}$ in iteration \textit{t} is obtained by maximizing the natural logarithm of the posterior distribution of $\vec{u}_{i_k=n}$ given by:
\begin{equation}
\begin{array}{l}
H(\vec{u}_{i_k=n}):=\max_{\vec{u}_{i_k=n}}\Bigl[
\sum_{i \in I_t: i_k = n} [\frac{\phi_i}{2} - \frac{\omega_i\phi_i^{2}}{2}]\\
 \qquad \ \- \frac{(\vec{u}_{i_k=n} \oslash \vec{\mu}_{i_k=n})^\top (\vec{u}_{i_k = n} \oslash \vec{\mu}_{i_k = n})}{2}
\Bigr],
\end{array}
\label{fUeq:eqn}
\end{equation}
where the operator $\oslash$ represents element-wise division between the two vectors $\vec{u}_{i_k=n}$ and $\vec{\mu}_{i_k = n}$.

Equation (\ref{fUeq:eqn}) is a quadratic equation in $\vec{u}_{i_k=n}$ and hence has a \emph{closed-form} update.

Subsequently the update for $\mu_{i_k=n,r}^{2}$ at time step $t$ is given by:
\begin{equation}\label{eq:update_mu_without_target}
{\mu_{i_k=n,r}^{2}}_{(t)} = \frac{{u_{i_k=n,r}^{2}}_{(t-1)}}{2a_c + 3} + \frac{2b_2}{2a_c + 3}.
\end{equation}

\textbf{Mode $k$ with binary target information}:
Given binary target information for mode $k$ we drop the notation $(k)$ for all variables. Let $z_{l,n}$ denote the binary label (either $+1$ or $-1$) for element $n$ for task $l$. Let $\nu_{l,n}$ correspond to the data augmented variable that is P\'{o}lya-Gamma distributed for element the $n$ and task $l$.

The update for $\vec{u}_{i_k=n}$ in iteration \textit{t} is obtained by maximizing the natural logarithm of the posterior distribution of $\vec{u}_{i_k=n}$ given by:
\begin{equation}
\begin{array}{ll}
H(\vec{u}_{i_k=n}):=\max_{\vec{u}_{i_k=n}}\Bigl[
\sum_{i \in I_t: i_k = n} [\frac{\phi_i}{2} - \frac{\omega_i\phi_i^{2}}{2}] \\
\qquad - \frac{(\vec{u}_{i_k=n} \oslash \vec{\mu}_{i_k=n})^\top (\vec{u}_{i_k=n} \oslash \vec{\mu}_{i_k = n})}{2} \\
\qquad + \sum_{l = 1}^{L} [z_{l,n}\frac{\vec{u}_{i_k=n}^{\top}\vec{\hat{\beta}_l}}{2} - \frac{\hat{\nu}_{l,n}[\beta_{0,l} + \vec{u}_{i_k=n}^{\top}\vec{\hat{\beta}}_l]^{2}}{2}]
\Bigr],
\end{array}
\label{fUeqSI:eqn}
\end{equation}
where $\vec{\hat{\beta}}_l$ denotes the vector of $R$ coefficients without the bias.
Note that in (\ref{fUeqSI:eqn}) we are training a Logistic model using the binary target information to detect abusive entities. And the operator $\oslash$ represents element-wise division between the two vectors $\vec{u}_{i_k=n}$ and $\vec{\mu}_{i_k = n}$. Also (\ref{fUeqSI:eqn}) is a quadratic equation in $\vec{u}_{i_k=n}$ and hence has a \emph{closed-form} update.

Subsequently the update for $\mu_{i_k=n,r}^{2}$ at time step $t$ is given by:
\begin{equation}\label{eq:update_mu_with_target}
{\mu_{i_k=n,r}^{2}}_{(t)} = \frac{{u_{i_k=n,r}^{2}}_{(t-1)}}{2a_c + 3} + \frac{2b_1}{2a_c + 3}.
\end{equation}

\subsection*{Algorithm}
Algorithm $1$ presents the pseudo-code for the multi-target semi-supervised CP tensor decomposition using partial natural gradients.

\begin{algorithm}[h]
    \caption{Partial Natural Gradient.}
    \label{alg:naturalGrads}
    \small
    \begin{enumerate}
\item Randomly initialize $\vec{\tau}, \vec{\lambda}$, $\vec{u}^{(1)}_{1} \ldots \vec{u}^{(K)}_{n_K}$ and $\vec{\beta}_l^{(1)} \cdots \vec{\beta}_L^{(K)}$.
\item Set the step-size schedule $\gamma_t$ appropriately.
\item \textbf{repeat}
\begin{enumerate}
\item Sample (with replacement) mini-batch $I_t$ from the training data.
\item For $i \in I_t$ set \\
--\ \ \ \ \ \  $A_i^{r} = \prod_{k=1}^K u_{i_k,r}^{(k)}$ for $r \in [1,R]$ \\
-- \ \ \ \ \ \ $\hat{\omega}_i = \frac{\text{tanh}(\frac{\phi_i}{2})}{2\phi_i}$ where $\phi_i = \vec{\lambda}^{\top}A_i$ \\
-- \ \ \ \ \ \ $N_{ii} = \frac{1}{\Bigl[\text{exp}[-\frac{\phi_i}{2}] + \text{exp}[\frac{\phi_i}{2}] \Bigr]^2}$.
\item For $k \in [1, K]$ and $l \in [1,L]$ (where applicable) set \\
-- \ \ \ \ \ \ $C_{i_k=n,r}^{(k)} = \lambda_r\prod_{k' \ne k}^K u_{i_{k'},r}^{(k')}$ for $r \in [1,R]$ \\
-- \ \ \ \ \ \ For entity $i_k=m$ with binary target information $z_{l,m}^{(k)}$:\\
\ \ \ \ \ \ \ \ \ \ $\hat{\nu}_{l,m}^{(k)} = \frac{\text{tanh}(\frac{\psi_{l,m}^{(k)}}{2})}{2\psi_{l,m}^{(k)}}$ where $\psi_{l,m}^{(k)} = \vec{\beta}_l{^{(k)}}^{\top}\tilde{\vec{u}}_{i_k=m}$\\
\ \ \ \ \ \ \ \ \ \ \ $O_{l:m=n} = \frac{1}{\Bigl[\text{exp}[-\frac{\psi_{l,m=n}}{2}] + \text{exp}[\frac{\psi_{l,m=n}}{2}] \Bigr]^2}$ \\
-- \ \ \ \ \ \ Compute \textit{gradient} and \textit{partial Fisher information matrix}\\
\ \ \ \ \ \ \ \ \ \ \ w.r.t. $\vec{\beta}_l^{(k)}$ and update $\vec{\beta}_l^{(k)}$ using (\ref{sgd:eqn}). \\
-- \ \ \ \ \ \ Update $\rho_{l,r}{^{(k)}}^2$ using (\ref{eq:update_rho}).\\
-- \ \ \ \ \ \ Compute \textit{gradient} and \textit{partial Fisher information matrix}\\
\ \ \ \ \ \ \ \ \  w.r.t. $\vec{u}_{i_k:i \in I_t}^{(k)}$ and update $\vec{u}_{i_k:i \in I_t}^{(k)}$ using (\ref{sgd:eqn}). \\
-- \ \ \ \ \ \ Update $\mu_{i_k:i \in I_t,r}^2$ using (\ref{eq:update_mu_without_target}) or (\ref{eq:update_mu_with_target}).
\item  Compute \textit{gradient} and \textit{partial Fisher information matrix} w.r.t. $\vec{\lambda}$ \\
\ \ and update $\vec{\lambda}$ using (\ref{sgd:eqn}).
\item Update $\vec{\tau}$ using (\ref{tauEq:eqn}).
\end{enumerate}
\item \textbf{until} forever
\end{enumerate}
\end{algorithm}

\section{Computation of Partial Fisher information matrix} \label{supp_sec:fisher_matrix}

\subsection*{Partial Fisher information matrix w.r.t. $\vec{\lambda}$}

Consider the exponent of the function to be maximized w.r.t. $\vec{\lambda}$, which is:
{\small
\begin{equation}
g(\vec{\lambda}) = \exp\Bigl[\sum_{i \in I_t} \kappa_i \phi_i - \frac{\omega_i \phi_i^{2}}{2}\Bigr] \exp\Bigl[-\frac{(\vec{\lambda} \oslash \sqrt{\vec{\tau}})^\top (\vec{\lambda} \oslash \sqrt{\vec{\tau}})}{2}\Bigr],
\label{fMax:eqn}
\end{equation}
}where $\kappa_{i} = y_i - \frac{1}{2}$ and $y_i \in \{0, 1\}$. And the operator $\oslash$ represents element-wise division between the two vectors $\vec{\lambda}$ \& $\vec{\tau}$.

The first exponential term of the right hand side (RHS) of (\ref{fMax:eqn}) is the joint conditional likelihood of the binary outcome $y_i \in \{0, 1\}$, denoted as $P(y_i|\vec{\lambda}, A_i)$ and the conditional likelihood of the P\'olya-Gamma distributed variable (from data augmentation) denoted as $P(\omega_i|\vec{\lambda}, A_i)$. The second exponential term of the RHS of (\ref{fMax:eqn}) is the {\it Gaussian} prior on $\vec{\lambda}$ with variance $\vec{\tau}$.
The stochastic natural gradient ascent-style updates for $\vec{\lambda}$ at step $t$ is given as:
\begin{equation} 
\vec{\lambda}_{(t)} = \vec{\lambda}_{(t-1)} + \gamma_t.\mathcal{I}(\vec{\lambda}_{(t-1)})^{-1}\mathop{\mathbb{E}}_{y_i, \omega_i: i \in I_t} \Bigl[\nabla_{\vec{\lambda}} \log[g(\vec{\lambda})]\Bigr], \label{sgd2:eqn}
\end{equation}
where $\gamma_t$ in (\ref{sgd2:eqn}) is given by the following equation:
\begin{equation*}
\gamma_t = \frac{1}{(\tau_{p} + t)^{\theta}}.
\end{equation*}
$\mathcal{I}(\vec{\lambda}_{(t-1)})^{-1}$ in (\ref{sgd2:eqn}) denotes the inverse of the partial Fisher information matrix whose size is $R \times R$ since the second order derivatives are computed only w.r.t. $\vec{\lambda}$.

Note that the joint likelihood term in (\ref{fMax:eqn}) is un-normalized. The partial Fisher information matrix is computed only w.r.t. the data i.e., $y_i$. To do this, we first compute the expectation w.r.t. the data augmented variable $\omega_i$. The resulting equation is a Logistic function from the following identity:
\begin{equation}
\frac{\exp[\phi_i]^{y_i}}{1 + \exp[\phi_i]} = \frac{\exp[\kappa_i\phi_i]}{2}\int_0^{\infty} \exp[-\frac{\omega_i\phi_i^2}{2}]p(\omega_i)\mathrm{d}\omega_i.
\label{cosh:eqn}
\end{equation}
There is a closed-form solution for the integral in (\ref{cosh:eqn}), hence we obtain:
\begin{equation}
\frac{\exp[\phi_i]^{y_i}}{1 + \exp[\phi_i]} = \frac{\exp[\kappa_i\phi_i]}{2}\frac{1}{\mathrm{cosh}[\frac{\phi_i}{2}]}.
\label{cosh_2:eqn}
\end{equation}
Equation (\ref{cosh_2:eqn}) is a normalized likelihood for each $i \in I_t$ and let it be denoted by $\mathcal{L}_i$. Using the definition of hyperbolic cosine, (\ref{cosh_2:eqn}) becomes:
\begin{equation*}
\mathcal{L}_i = \frac{\exp \Bigl[\kappa_i\phi_i\Bigr]}{\exp \Bigl[-\frac{\phi_i}{2}\Bigr] + \exp \Bigl[\frac{\phi_i}{2}\Bigr]}.
\end{equation*}
To this end, the partial Fisher Information with respect to $\vec{\lambda}$ for $\mathcal{L}_i$ and $i \in I_t$ is given by:
\begin{equation*}
I_{\mathcal{L}}(\vec{\lambda}) = -\mathop{\mathbb{E}}_{y_i: i \in I_t} \Bigl[\sum_{i \in I_t} \frac{\partial^{2} log[\mathcal L_i]}{\partial \vec{\lambda}^{2}}\Bigr] = [\boldsymbol{A}_{I_t}^\top  \boldsymbol{N}_{I_t} \boldsymbol{A}_{I_t}],
\end{equation*}
where $\boldsymbol{A}_{I_t}$ denotes the matrix whose rows are $A_i$ for $i \in I_t$ and $\boldsymbol{N}_{I_t}$ denotes the diagonal matrix whose diagonal elements are $N_{ii}$; where:
\begin{equation*}
N_{ii} = \frac{1}{\Bigl[\exp[-\frac{\phi_i}{2}] + \exp [\frac{\phi_i}{2}]\Bigr]^2}.
\end{equation*}
The prior term is accounted by considering its variance as a conditioner, hence the conditioned partial Fisher Information matrix for the parameter $\vec{\lambda}$ at step $t$ is given by:
\begin{equation}\label{eq:fisher_information_lambda}
\mathcal{I}(\vec{\lambda}_{(t-1)}) = [\boldsymbol{A}_{I_t}^\top  \boldsymbol{N}_{I_t} \boldsymbol{A}_{I_t}] + \text{diag}[\vec{\tau}_{(t-1)}]^{-1},
\end{equation}
where diag$[\vec{\tau}_{(t-1)}]^{-1}$ denotes inverse of a diagonal matrix whose diagonal is $\vec{\tau}_{(t-1)}$.

\subsection*{Partial Fisher information matrix w.r.t. $\vec{\beta}_{l}^{(k)}$ for task $l$ in mode $k$}

The function to be maximized w.r.t. $\vec{\beta}_{l}^{(k)}$ is:
\begin{equation}
\begin{split}
F(\vec{\beta}_{l}^{(k)})=\Bigl[
\sum_{m = 1}^{M} [z_{l,m}^{(k)}\frac{\psi_{l,m}^{(k)}}{2} - \frac{\nu_{l,m}^{(k)} \psi_{l,m}{^{(k)}}^{2}}{2}]
- \frac{(\vec{\beta}_{l}^{(k)} \oslash \vec{\rho}_{l}^{(k)})^\top (\vec{\beta}_{l}^{(k)} \oslash \vec{\rho}_{l}^{(k)})}{2}]
\Bigr],
\end{split}
\label{fBeta:eqn}
\end{equation}
where the operator $\oslash$ represents element-wise division between the two vectors $\vec{\beta}_{l}^{(k)}$ \& $\vec{\rho}_{l}^{(k)}$.

Let $\tilde{\boldsymbol{U}}_M^{(k)}$ be the matrix whose rows are $\tilde{\vec{u}}_{i_k=m}^{(k)}$ for $m \in [1,M]$ for mode $k$ with target information.
We compute the partial Fisher information matrix $I(\vec{\beta}{_{l}^{(k)}}_{(t-1)})$ similar to $\vec{\lambda}$ as:
\begin{equation*}
\mathcal{I}(\vec{\beta}_{l}^{(k)}) = [\tilde{\boldsymbol{U}}_{M}^{(k)}]^\top \boldsymbol{O}_{l,M} \tilde{\boldsymbol{U}}_{M}^{(k)} + \text{diag}[\vec{\rho}_{l}{^{(k)}}^{2}]^{-1},
\end{equation*}
where the $\boldsymbol{O}_{l,M}$ is the diagonal matrix whose diagonal elements are $O_{l:m,m}$ where:
\begin{equation*}
O_{l:m,m} = \frac{1}{\Bigl[\exp[-\frac{\psi_{l,m}^{(k)}}{2}] + \exp [\frac{\psi_{l,m}^{(k)}}{2}]\Bigr]^2}.
\end{equation*}

\subsection*{Partial Fisher information matrix w.r.t. $\vec{u}_{i_k=n}^{(k)}$ for mode $k$: Without Target Information}

The function to be maximized w.r.t. factor $\vec{u}_{i_k=n}^{(k)}$ is:
\begin{equation}
\begin{array}{l}
H(\vec{u}_{i_k=n}^{(k)})=\Bigl[ 
\sum_{i \in I_t: i_k = n} [\frac{\phi_i}{2} - \frac{\omega_i \phi_i^{2}}{2}]
- \frac{(\vec{u}_{i_k=n}^{(k)} \oslash \vec{\mu}_{i_k=n})^\top (\vec{u}_{i_k=n}^{(k)} \oslash \vec{\mu}_{i_k = n})}{2}
\Bigr],
\end{array}
\label{fUeq:eqn}
\end{equation}
where the operator $\oslash$ represents element-wise division between the two vectors $\vec{u}_{i_k=n}^{(k)}$ \& $\vec{\mu}_{i_k = n}$.

Similarly we can compute the partial Fisher information for $\vec{u}_{i_k=n}^{(k)}$ corresponding to element $n$ in mode $k$ as:
\begin{equation*}
\mathcal{I}(\vec{u}_{i_k=n}^{(k)}{_{(t-1)}}) = [\boldsymbol{C}_{n}^{(k)}]^\top \boldsymbol{N}_{n} \boldsymbol{C}_{n}^{(k)} + \text{diag}[\vec{\mu}_{i_k=n}^{2}]^{-1},
\end{equation*}
where $\boldsymbol{C}_n^{(k)}$ is a matrix whose rows are $\vec{C}_{i_k = n}$ for $i \in I_t$ and $\boldsymbol{N}_{n}$ is the diagonal matrix whose diagonal elements $N_{i:i_k = n}$ is given by:
\begin{equation*}
N_{i:i_k=n} = \frac{1}{\Bigl[\exp[-\frac{\phi_{i:i_k=n}}{2}] + \exp[\frac{\phi_{i:i_k=n}}{2}]\Bigr]^2}.
\end{equation*}

\subsection*{Partial Fisher information matrix w.r.t. $\vec{u}_{i_k=n}^{(k)}$ for mode $k$: With Binary Target Information}

The function to be maximized w.r.t. factor $\vec{u}_{i_k=n}^{(k)}$ is:
\begin{equation}
\begin{array}{lll}
H(\vec{u}_{i_k=n}^{(k)}) & = & \Bigl[
\sum_{i \in I_t: i_k = n} [\frac{\phi_i}{2} - \frac{\omega_i \phi_i^{2}}{2}] 
 - \frac{(\vec{u}_{i_k=n}^{(k)} \oslash \vec{\mu}_{i_k=n})^\top (\vec{u}_{i_k=n}^{(k)} \oslash \vec{\mu}_{i_k = n})}{2}\\
&  & \quad + \sum_{l=1}^L [z_{l,n}^{(k)}.\frac{(\vec{u}_{i_k=n}^{(k)})^{\top}\vec{\hat{\beta}}_l^{(k)}}{2} - \frac{\nu_{l,n}^{(k)}[\beta_{0,l}^{(k)} + (\vec{u}_{i_k=n}^{(k)})^{\top}\vec{\hat{\beta}}_l^{(k)}]^{2}}{2}]
\Bigr],
\end{array}
\label{fUeqSI:eqn}
\end{equation}
where the operator $\oslash$ represents element-wise division between the two vectors $\vec{u}_{i_k=n}^{(k)}$ \& $\vec{\mu}_{i_k = n}$. Note that (\ref{fUeqSI:eqn}) does not have the non-negativity constraint since we are training a Logistic model using the binary target information to detect abusive entities.

Similarly we can compute the partial Fisher information for $\vec{u}_{i_k=n}^{(k)}$ corresponding to element $n$ in mode $k$ as:
{\small
\begin{equation*}
\begin{array}{l}
\mathcal{I}(\vec{u}_{i_k=n}^{(k)}{_{(t-1)}}) = [\boldsymbol{C}_{n}^{(k)}]^\top \boldsymbol{N}_{n} \boldsymbol{C}_{n}^{(k)} + \sum_{l=1}^L [\vec{\beta}_l^{(k)}]^\top O_{l:m=n} \vec{\beta}_l^{(k)}  +  \text{diag}[\vec{\mu}_{i_k=n}^{2}]^{-1},
\end{array}
\end{equation*}}
where $O_{l:m=n}$ is:
\begin{equation*}
O_{l:m=n} = \frac{1}{\Bigl[\exp[-\frac{\psi_{l,m=n}^{(k)}}{2}] + \exp[\frac{\psi_{l,m=n}^{(k)}}{2}]\Bigr]^2}.
\end{equation*}

\section{Computation of the gradient}\label{supp_sec:gradient}

\subsection*{Gradient w.r.t. $\vec{\lambda}$}
To update $\vec{\lambda}$, we compute the gradient of the natural logarithm of (\ref{fMax:eqn}) as:
{\small
\begin{equation}\label{grad_lambda:eqn}
\begin{array}{l}
\mathop{\mathbb{E}}_{y_i, \omega_i: i \in I_t} \Bigl[\nabla_{\vec{\lambda}} \log[g(\vec{\lambda})]\Bigr] = \mathop{\mathbb{E}}_{y_i, \omega_i: i \in I_t} \Bigl[\Bigl[\sum_{i \in I_t} \kappa_iA_i - A^\top_i \omega_i (A_i \vec{\lambda}) \Bigr] - \text{diag}[\vec{\lambda} \oslash \vec{\tau}] \Bigr].
\end{array}
\end{equation}
}

Separating out terms independent of $y_i$ \& $\omega_i$ and replacing with matrix operations where applicable; the RHS of (\ref{grad_lambda:eqn}) becomes:
\begin{equation*}
\sum_{i \in I_t} \mathop{\mathbb{E}}_{y_i} \Bigl[\kappa_i \Bigr]A_i - \Bigl[\boldsymbol{A}_{I_t}^\top \hat{\boldsymbol{\omega}}_{I_t} \boldsymbol{A}_{I_t}  + \text{diag}[\vec{\tau}]^{-1} \Bigr]\vec{\lambda},
\end{equation*}
where $\hat{\boldsymbol{\omega}}_{I_t}$ is a diagonal matrix whose diagonal elements are $\hat{\omega}_i$, where:
\begin{equation*}
\hat{\omega}_i = \mathop{\mathbb{E}}[\omega_i] = \frac{\text{tanh}(\frac{\phi_i}{2})}{2\phi_i}.
\end{equation*}

Now $\kappa_i = -0.5$ if  $y_i = 0$ and $\kappa_i = 0.5$ if $y_i = 1$. Hence:
\begin{equation*}
\begin{array}{l}
\mathop{\mathbb{E}}_{y_i} \Bigl[\kappa_i | y_i = 0  \Bigr] = -0.5\frac{\exp(-\frac{\phi_i}{2})}{\exp(-\frac{\phi_i}{2}) + \exp(\frac{\phi_i}{2})} \quad \text{and}\\
\mathop{\mathbb{E}}_{y_i} \Bigl[\kappa_i | y_i = 1  \Bigr] = 0.5\frac{\exp(\frac{\phi_i}{2})}{\exp(-\frac{\phi_i}{2}) + \exp(\frac{\phi_i}{2})}.
\end{array}
\end{equation*}

\subsection*{Gradient w.r.t. $\vec{\beta}_l^{(k)}$}
To update $\vec{\beta}_l^{(k)}$, we compute the gradient of (\ref{fBeta:eqn}) as:
\small{
\begin{equation}\label{grad_beta:eqn}
\begin{array}{lll}
\mathop{\mathbb{E}}_{z_{l,m}^{(k)}, \nu_{l,m}^{(k)}: m \in [1, M]} \Bigl[\nabla_{\vec{\beta}_l^{(k)}} F(\vec{\beta}_l^{(k)})\Bigr] & = & \mathop{\mathbb{E}}_{z_{l,m}^{(k)}, \nu_{l,m}^{(k)}: m \in [1,M]} \Bigl[ 
\sum_{m = 1}^{M} [z_{l,m}^{(k)} \frac{\tilde{\vec{u}}^{(k)}_{i_k = m}}{2}  \\
 & & - ({\tilde{\vec{u}}^{(k)}_{i_k = m}}^\top \nu_{l,m}^{(k)} \tilde{\vec{u}}^{(k)}_{i_k = m})\vec{\beta}_l^{(k)}]
- \text{diag}[\vec{\beta}_l^{(k)} \oslash \vec{\rho}_l]
\Bigr].
\end{array}
\end{equation}}Separating out terms independent of $z_{l,m}^{(k)}$ \& $\nu_{l,m}^{(k)}$ and replacing with matrix operations where applicable; the RHS of (\ref{grad_beta:eqn}) becomes:
\begin{equation*}
\sum_{m = 1}^{M} \mathop{\mathbb{E}}_{z_{l,m}^{(k)}} \Bigl[z_{m}^{(k)} \Bigr]\frac{\tilde{\vec{u}}^{(k)}_{i_k = m}}{2} - \Bigl[{\tilde{\boldsymbol{U}}_{M}^{(k)\top}}  \hat{\boldsymbol{\nu}}_{l,M}^{(k)} \tilde{\boldsymbol{U}}_{M}^{(k)} + \text{diag}[\vec{\rho}{^{(k)}}^2]^{-1} \Bigr]\vec{\beta}_l^{(k)},
\end{equation*}
where $\hat{\boldsymbol{\nu}}_{l,M}^{(k)}$ is a diagonal matrix whose diagonal elements are $\hat{\nu}_{l,m}^{(k)}$, where:

Subsequently:
\begin{equation*}
\begin{array}{l}
\mathop{\mathbb{E}}_{z_{l,m}^{(k)}} \Bigl[z_{l,m}^{(k)} | label_{m}^{(k)} = 0  \Bigr] = -\frac{\exp(-\frac{\psi_{l,m}^{(k)}}{2})}{\exp(-\frac{\psi_{l,m}^{(k)}}{2}) + \exp(\frac{\psi_{l,m}^{(k)}}{2})} \quad \text{and}\\
\mathop{\mathbb{E}}_{z_{l,m}^{(k)}} \Bigl[z_{l,m}^{(k)} | label_{m}^{(k)} = 1  \Bigr] = \frac{\exp(\frac{\psi_{l,m}^{(k)}}{2})}{\exp(-\frac{\psi_{l,m}^{(k)}}{2}) + \exp(\frac{\psi_{l,m}^{(k)}}{2})}.
\end{array}
\end{equation*}

\subsection*{Gradient w.r.t. $\vec{u}_{i_k=n}^{(k)}$: Without Target Information}
To update $\vec{u}_{i_k=n}^{(k)}$, we compute the gradient of (\ref{fUeq:eqn}) as:
\begin{equation*}
\sum_{i \in I_t:i_k=n} \mathop{\mathbb{E}}_{y_i} \Bigl[\kappa_i \Bigr]C_{i_k=n}^{(k)} - \Bigl[ {\boldsymbol{C}_{n}^{(k)}}^\top \hat{\boldsymbol{\omega}}_n \boldsymbol{C}_{n}^{(k)}  + \text{diag}[\vec{\mu}_{i_k=n}^{2}]^{-1} \Bigr]\vec{u}_{i_k=n}^{(k)},
\end{equation*}
where $\hat{\boldsymbol{\omega}}_{n}$ is a diagonal matrix whose diagonal elements are $\hat{\omega}_{i \in I_t:i_k=n}$.

\subsection*{Gradient w.r.t. $\vec{u}_{i_k=n}^{(k)}$: With Binary Target Information}

To update $\vec{u}_{i_k=n}^{(k)}$, we compute the gradient of (\ref{fUeqSI:eqn}) as:
\begin{equation*}
\begin{array}{lll}
\sum_{i \in I_t:i_k=n} \mathop{\mathbb{E}}_{y_i} \Bigl[\kappa_i \Bigr]C_{i_k=n}^{(k)} + \sum_{l=1}^L \mathop{\mathbb{E}}_{z_{l,m=n}^{(k)}} \Bigl[z_{l,m=n}^{(k)} \Bigr]\frac{\hat{\vec{\beta}}_l^{(k)}}{2} 
- \Bigl[{\boldsymbol{C}_{n}^{(k)}}^\top \hat{\boldsymbol{\omega}}_n \boldsymbol{C}_{n}^{(k)}  + \text{diag}[\vec{\mu}_{i_k=n}^{2}]^{-1} - \sum_{l=1}^L ({\hat{\vec{\beta}}_l^{(k)}}^\top\hat{\nu}_{l,m=n}^{(k)}\hat{\vec{\beta}}_l^{(k)}) \Bigr]\vec{u}_{i_k=n}^{(k)}.
\end{array}
\end{equation*}

\section{Confidence Interval for AUC}\label{supp_sec:CI}

\subsection*{Detection Abusive Sellers}
\begin{figure}[t]
\center
\includegraphics[scale=0.30]{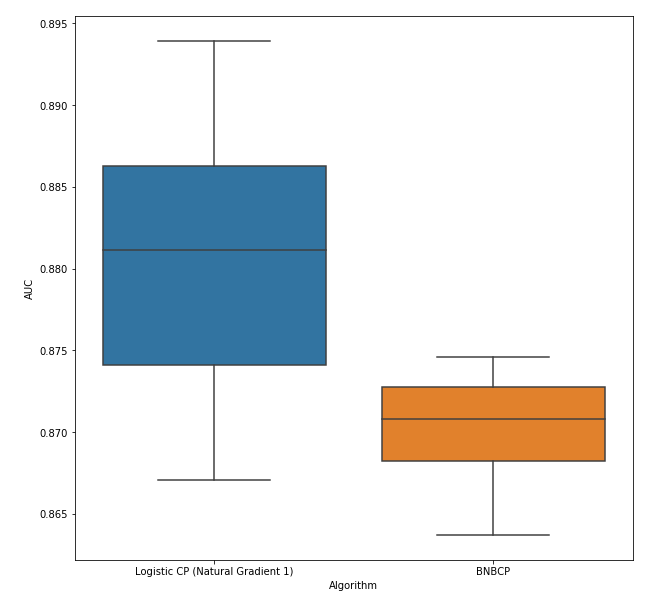}
\caption{\label{auc1:fig} AUC Box Plot: Detecting Abusive Sellers.}
\end{figure}

Figure~\ref{auc1:fig} shows the confidence intervals (box plot) of AUC across eleven runs of the $BNBCP$ and $Logistic$ $CP$ ($Natural$ $Gradient$ $1$) semi-supervised models on the $SELLER-TENSOR$ data.
 
\end{document}